\documentclass[10pt,twocolumn,letterpaper]{article}

\usepackage{cvpr}
\usepackage{times}
\usepackage{epsfig}
\usepackage{graphicx}
\usepackage{amsmath}
\usepackage{amssymb}
\usepackage{slashbox}
\usepackage{array,multirow}
\usepackage{booktabs}
\usepackage{comment}
\usepackage{rotating}
\usepackage{verbatim}
\usepackage{stackengine}
\usepackage{caption}
\usepackage[utf8]{inputenc}
\usepackage{color}
\usepackage[export]{adjustbox}
\usepackage{booktabs}
\usepackage{xr} 
\usepackage[breaklinks=true,bookmarks=false]{hyperref}

\def\cvprPaperID{2285} 

\cvprfinalcopy

\usepackage[dvipsnames,svgnames,x11names]{xcolor}

\newcolumntype{C}{>{\centering\arraybackslash}p{3em}}
\newcolumntype{D}{>{\centering\arraybackslash}p{6em}}

\begin{document}

\title{Improving Landmark Localization with Semi-Supervised Learning}

\author{Sina Honari${^1}$\thanks{Part of this work was done when author was at NVIDIA Research} \,, Pavlo Molchanov${^2}$, Stephen Tyree${^2}$, Pascal Vincent${^{1,4,5}}$, Christopher Pal${^{1,3}}$, Jan Kautz${^2}$\\
{\normalsize ${^1}$MILA-University of Montreal, ${^2}$NVIDIA, ${^3}$Ecole Polytechnique of Montreal, ${^4}$CIFAR, ${^5}$Facebook AI Research.}\\
{\tt\small ${^1}$\{honaris, vincentp\}@iro.umontreal.ca,} \\
{\tt\small ${^2}$\{pmolchanov, styree, jkautz\}@nvidia.com, ${^3}$christopher.pal@polymtl.ca}}

\maketitle


\begin{abstract}
We present two techniques to improve landmark localization in images from partially annotated datasets. Our primary goal is to leverage the common situation where precise landmark locations are only provided for a small data subset, but where class labels for classification or regression tasks related to the landmarks are more abundantly available. 
First, we propose the framework of sequential multitasking and explore it here through an architecture for landmark localization where training with class labels acts as an auxiliary signal to guide the landmark localization on unlabeled data. A key aspect of our approach is that errors can be backpropagated through a complete landmark localization model. Second, we propose and explore an unsupervised learning technique for landmark localization based on having a model predict equivariant landmarks with respect to transformations applied to the image. We show that these techniques, improve landmark prediction considerably and can learn effective detectors even when only a small fraction of the dataset has landmark labels. We present results on two toy datasets and four real datasets, with hands and faces, and report new state-of-the-art on two datasets in the wild, e.g. with only 5\% of labeled images we outperform previous state-of-the-art trained on the AFLW dataset. 
\end{abstract}
\section{Introduction}
\begin{figure}[t!]
\begin{center}
\resizebox{1\linewidth}{!}{
\begin{tabular}{cc}
\includegraphics[width=.438\textwidth, left]{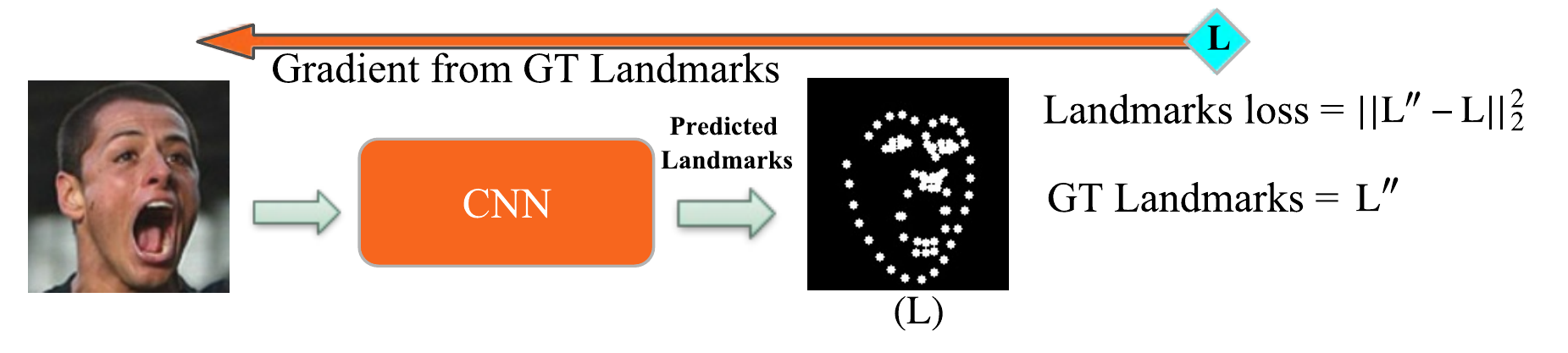} & S\\ 
\midrule
\includegraphics[width=.5\textwidth, left]{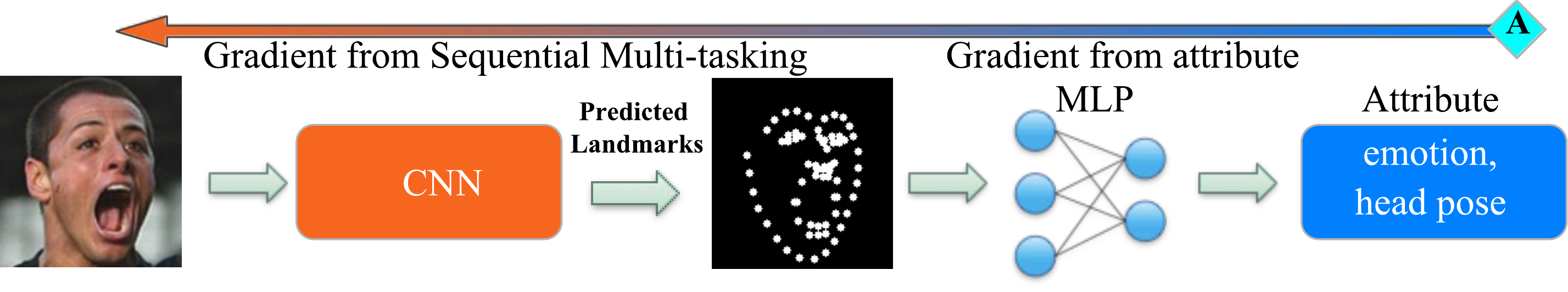} & M\\
\midrule
\includegraphics[width=.444\textwidth, left]{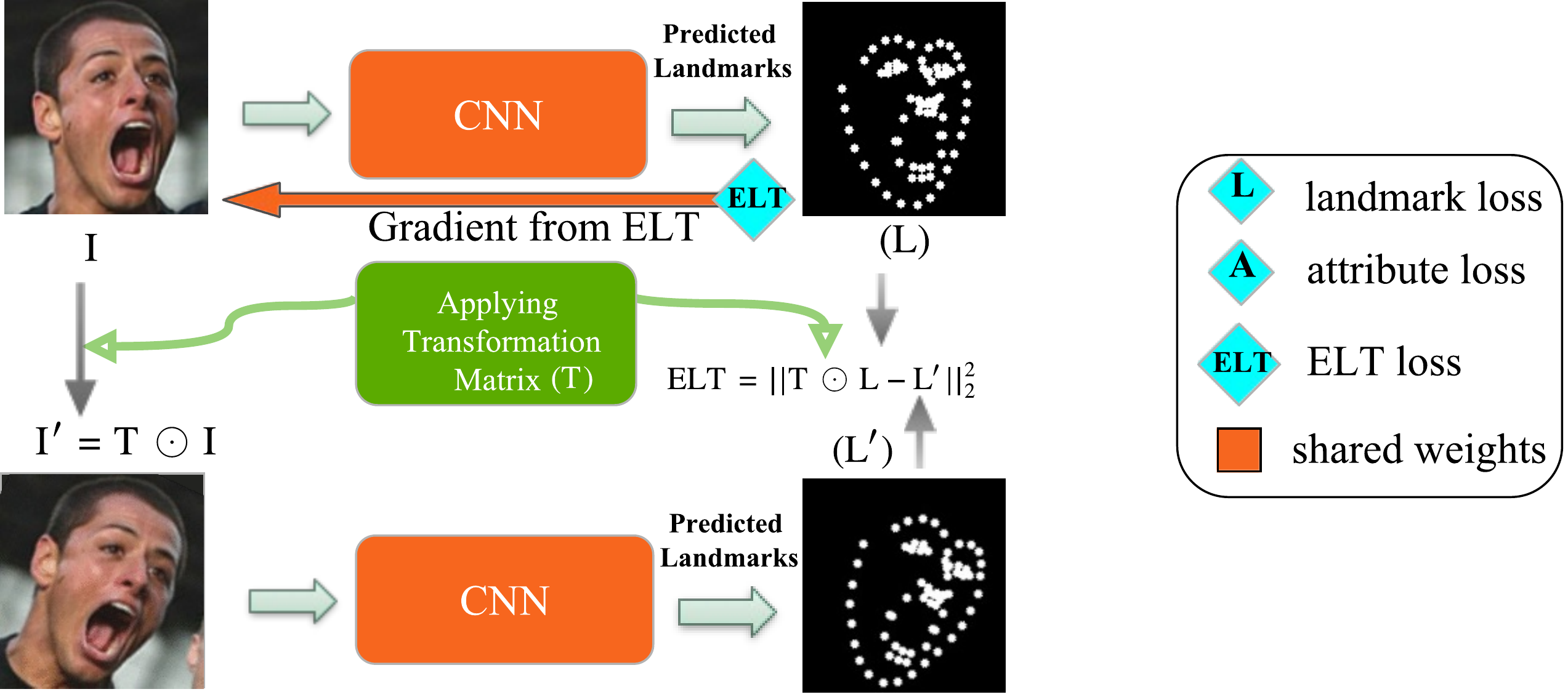} & N\\
\end{tabular}
}
\end{center}%
\vspace{-8pt}
\caption{\small In our approach three sources of gradients are used for learning a landmark localization network, from top to bottom: 
1) The gradient from $S$ labeled image-landmark pairs. 
2) The gradient from $M$ attribute examples, obtained through sequential multitasking. The first part of the network (a CNN) predicts landmarks with a soft-argmax output layer to make the entire network fully differentiable. The predicted landmarks (as $x$, $y$ pairs) are then fed into a multi-layer perceptron (MLP) for attribute regression/classification. 3) The gradient received from an unsupervised component of the composite loss which we refer to as an equivariant landmark transformation (ELT) (applied to $N$ images). This loss encourages the model to output landmarks that are equivariant to transformations applied to the image. Importantly, MLP regression and the ELT are applied to the model's predictions and not the ground truth (GT) landmarks, so they can be applied on images that are not labelled with landmarks. Our proposed approach allows efficient training even when $S \ll M \le N$.
}
\label{fig:both_archs}
\end{figure}

Landmark localization -- finding the precise location of specific parts in an image -- is a central step in many complex vision problems. Examples include hand tracking \cite{hu2010hand, datcu2013free},  gesture recognition \cite{dardas2010hand}, facial expression recognition \cite{kahou2013combining}, face identity verification \cite{taigman2014deepface, sun2014deep}, and eye gaze tracking \cite{zhang2015appearance, mora2012gaze}. 
Reliable landmark estimation 
is often part of the pipeline for sophisticated, robust vision tools.
Neural networks have yielded state-of-the art results on numerous landmark estimation problems \cite{tompson2015efficient, honari2016recombinator, xiao2016robust, yu2016deep, wang2016recurrent}. 
However, neural networks generally need to be trained on a large set of labeled data to be robust to the variations in natural images. Landmark labeling is a tedious manual work where precision is important; as a result, few landmark datasets are large enough to train reliable deep neural networks. On the other hand it is much easier to label an image with a single class label rather than the entire set of precise landmarks, and datasets with labels related to---but distinct from---landmark detection are far more abundant.

The key elements of our approach are illustrated in Figure \ref{fig:both_archs}. The top diagram illustrates a traditional convolutional neural network (CNN) based landmark localization network.
The first key element of our work -- illustrated in the second diagram of Figure \ref{fig:both_archs}, is that we use the indirect supervision of class labels to guide classifiers trained to localize landmarks. The class label can be considered a weak label that sends indirect signals about landmarks. For example, a photo of a hand gesture with the label ``waving'' likely indicates that the hand is posed with an open palm and spread fingers, signaling a set of reasonable locations for landmarks on the hand. We leverage class labels that are more abundant or more easily obtainable than landmark labels, putting our proposed method in the category of multi-task learning. A common approach \cite{zhang2014facialMTFL, zhang2016MAFL, zhang2014improving, devries2014multi} to multi-task learning uses a traditional CNN, in which a final common fully-connected (FC) layer feeds into separate branches, each dedicated to the output for a different task. This approach learns shared low-level features across the set of tasks and acts as a regularizer, particularly when the individual tasks have few labeled samples.

There is a fundamental caveat to applying such an approach directly to simultaneous classification and landmark localization tasks, because the two have opposing requirements: classification output needs to be insensitive (invariant) to small deformations such as translations, whereas landmark localization needs to be equivariant to them, i.e., follow them precisely with high sensitivity. To build in invariance, traditional convolutional neural networks for classification problems rely on pooling layers to integrate signals across the input image. However, tasks such as landmark localization or image segmentation require both the global integration of information as well as an ability to retain local, pixel-level details for precise localization.
The goal of producing precise landmark localization has thus led to the development of new layers and network architectures such as dilated convolutions \cite{yu2015multi}, stacked what-where auto-encoders \cite{Zhao2015StackedWA}, recombinator-networks \cite{honari2016recombinator}, fully-convolutional networks \cite{long2015fully}, and hyper-columns \cite{hariharan2015hypercolumns}, each preserving pixel-level information.
These models have however not been developed with multi-tasking in mind.

Current multi-task architectures \cite{zhang2014facialMTFL, zhang2016MAFL, zhang2014improving, devries2014multi, ranjan2016hyperface} predict landmark locations and auxiliary tasks as separate branches, i.e., \emph{in parallel}. In this scenario the auxiliary task is used for partial supervision of landmark localization model.
We propose a novel class of neural architectures which force classification predictions to flow through the intermediate step of landmark localization to provide complete supervision during backpropagation.

\textbf{
One of the contributions of our model is to
leverage auxiliary classification tasks and data, enhancing landmark localization by backpropagating classification errors through the landmark localization layers of the model.} 
Specifically, we propose a sequential architecture in which the first part of the network predicts landmarks via pixel-level heatmaps, maintaining high-resolution feature maps by omitting pooling layers and strided convolutions. The second part of the network computes class labels using predicted landmark locations. To make the whole network differentiable, we use soft-argmax for extracting landmark locations from pixel-level predictions.
Under this model, learning the landmark localizer is more directly influenced by the task of predicting class labels, allowing the classification task to enhance landmark localization learning.

Semi-supervised learning techniques \cite{salimans2016improved, rasmus2015semi, weston2012deep, wan2017crossing} have been used in deep learning to improve classification accuracy with a limited amount of labeled training data. A recently proposed method \cite{laine2016temporal} 
enforces invariance in class predictions over time and across a variety of data augmentations applied to unlabeled training data. \textbf{Our second contribution is to propose and explore an unsupervised learning technique for landmark localization where the model is asked to produce landmark localizations equivariant with respect to a set of transformations applied to the image.} In other words, we transform an image during training and ask the model to produce landmarks that are similarly transformed. Importantly, this technique does not require the true landmark locations, and thus can be applied during semi-supervised training to leverage images with unlabeled landmarks. This element of our work is illustrated in the third diagram of Figure \ref{fig:both_archs}. Independently from our work, Thewlis et al. \cite{Thewlis_2017_ICCV} 
proposed an unsupervised technique for landmark localization, however, the question if it can be used to improve supervised training remains open. 

To summarize, in this paper we make the following contributions:
1) We propose a novel multi-tasking neural architecture, which a) predicts landmarks as an intermediate step before classification in order to use the class labels to improve landmark localization, b) uses soft-argmax for a fully-differentiable model in which end-to-end training can be performed, even from examples that do not provide labeled landmarks.
2) We propose an unsupervised learning technique to learn features that are equivariant with respect to transformations applied to the input image. 
Combining contributions 1) and 2), we propose a robust landmark estimation technique which learns effective landmark predictors while requiring fewer labeled landmarks compared to current approaches.
3) We report state-of-the-art on 300W \cite{sagonas2013300} and AFLW \cite{AFLW_2011} without leveraging any external data.
 
\section{Sequential Multi-Tasking}
\label{sec:model_seq}
We refer to the new architecture that we propose for leveraging the attributes to guide the learning of landmark locations as \emph{sequential multi-tasking}. This architecture first predicts the landmark locations and then uses the predicted landmarks as the input to the second part of the network, which performs classification (see Fig.~\ref{fig:both_archs}-middle). In doing so, we create a bottleneck in the network, forcing it to solve the classification task only through the landmarks. If the goal were to enhance classification, this architecture would have been harmful since such bottlenecks \cite{he2016deep} would hurt the flow of information for classification. However, since our goal is landmark localization, this architecture enforces receiving signal from class labels through back-propagation to enhance landmark locations. 
This architecture benefits from auxiliary tasks that can be efficiently solved relying only on extracted landmark locations without observing the input image.

In order to make the whole pipeline trainable end-to-end, even on examples that do not provide any landmarks, we apply soft-argmax \cite{chapelle2010gradient} on the output of the last convolutional layer in the landmark prediction model. Specifically, let $M(I)$ be the stack of $K$ two-dimensional output maps produced by the last convolutional layer for a given network input image $I$. The map associated to the $k^{th}$ landmark will be denoted $M_k(I)$. To obtain a single 2d location $L_k=(x,y)$ for the landmark from $M_k(I)$, we use the following soft-argmax operation:
\begin{eqnarray}
L_k(I) &=& \textrm{soft-argmax}(\beta M_k(I)) \nonumber \\
&=& \sum_{i,j} \mathrm{softmax}(\beta M_k(I))_{i,j} (i,j) 
\end{eqnarray}
where softmax denotes a spatial softmax of the map, i.e. $\mathrm{softmax}(A)_{i,j} = {\exp(A_{i,j})}/{\sum_{i',j'} \exp(A_{i',j'})}$. $\beta$ controls the temperature of the resulting probability map, and $(i,j)$ iterate over pixel coordinates. 
In short, soft-argmax computes landmark coordinates $L_k = (x,y)$ as a weighted average of all pixel coordinate pairs $(i,j)$ where the weights are given by a softmax of landmark map $M_k$.

Predicted landmark coordinates are then fed into the second part of the network for attribute estimation. 
Having either classification or regression task, the model optimizes
\begin{eqnarray}
Cost\_attr=
\left\{
	\begin{array}{ll}
     - \log P(\mathbb{A} = \tilde{a} | {\mathbb{I}}  = {I}) &\text{, \ if classification}\\
     | \tilde{a} - a(I) | & \text{, \ if regression} \nonumber
	\end{array}
\right.
\end{eqnarray}
$P(\mathbb{A} = \tilde{a} | {\mathbb{I}}  = {I})$ denotes the probability ascribed by the model to the class $\tilde{a}$ given input image $I$, as computed by the final classification softmax layer. $\tilde{a}$ denotes the ground truth (GT) and $a(I)$ the predicted attributes in the regression task. 
Using soft-argmax, as opposed to a simple softmax, the model is fully differentiable through its landmark locations and is trainable end-to-end. 
\section{Equivariant Landmark Transformation}
We propose the following unsupervised learning technique to make the model's prediction consistent with respect to different transformations that are applied to the image. Consider an input image $I$ and the corresponding landmarks $L(I)$ predicted by the network. Now consider a small affine coordinate transformation $T$. We will use $T \odot \ldots$ to denote the application of such a transformation in coordinate space, whether it is applied to deform a bitmap image or to transform actual coordinates. If we apply this transformation to produce a deformed image $I'=T \odot I$ and compute the resulting landmark coordinates $L(I')$ predicted by the network, they should be very close to the result of applying the transformation on landmark coordinates $L(I)$, i.e., we expect to have $L(T\odot I) \approx T \odot L(I)$. The architecture for this technique, which we call \emph{equivariant landmark transformation (ELT)}, is illustrated in Fig.~\ref{fig:both_archs}-bottom. Multiple instances of $C_T$ can thus be added to the overall training cost, each corresponding to a different transformation $T$.

Our entire model is trained end-to-end to minimize the following cost
\begin{eqnarray}
&Cost = \quad \frac{1}{N} \sum_{(I, \tilde{a}) \in \mathcal{D}} \{ Cost\_attr \, + \nonumber \\
& \frac{\alpha}{K} \sum_{k=1}^{K} \| \underbrace{T \odot L_k(I)}_{\hat{L}_k} - \underbrace{L_k(T \odot I}_{L'_k}) \|_{2}^{2} \} + \nonumber \\
& \frac{\lambda}{S K} \sum_{\tilde{L}}  \sum_{k=1}^{K} || \tilde{L}_k - L_k(I) ||_{2}^{2} + \gamma || \mathbb{W} ||_{2}^2,
\end{eqnarray}
where $\mathcal{D}$ is the training set containing $N$ pairs $(I, \tilde{a})$ of input image and GT attribute. $K$ is the number of landmarks. $\tilde{L}_k$, $L_k(I)$ and $S$ respectively correspond to the GT, predicted landmarks and the number of images in the train set with labelled landmarks.
$\mathbb{W}$ represents the parameters of the model.
$\alpha$, $\lambda$, and $\gamma$ are weights for losses.
The first part of the cost is attribute classification or regression and affects the entire network. 
The second part is the ELT cost and can be applied to any training image, regardless of whether or not it is labeled with landmarks. This cost only affects the first part of the network (Landmark Localization).
  The third part is the squared Euclidean distance between GT and estimated landmark locations and is used only when landmark labels are provided. This cost only affects the first part of the network. The last cost is $\ell_2$-norm on the model's parameters.

\section{Experiments}
To validate our proposed model, we begin with two toy datasets in Sections \ref{sec:exp_shapes} and \ref{sec:exp_blocks}, in order to verify to what extent the class labels can be used to guide the landmark localization regardless of the complexity of the dataset. Later, we evaluate the proposed network on four real datasets: Polish sign-language dataset \cite{kawulok2014PolishHands} in Section \ref{sec:exp_hands}, Multi-PIE \cite{gross2010multi} in Section \ref{sec:exp_MultiPIE}, and two datasets in the wild; 300W~\cite{sagonas2013300} and AFLW~\cite{AFLW_2011} in Sections \ref{sec:exp_300W} and \ref{sec:exp_AFLW}. All the models are implemented in Theano \cite{2016arXiv160502688short}.
\subsection{Shapes dataset}
\label{sec:exp_shapes}
\begin{figure}
\begin{center}
\includegraphics[width=0.48\textwidth]{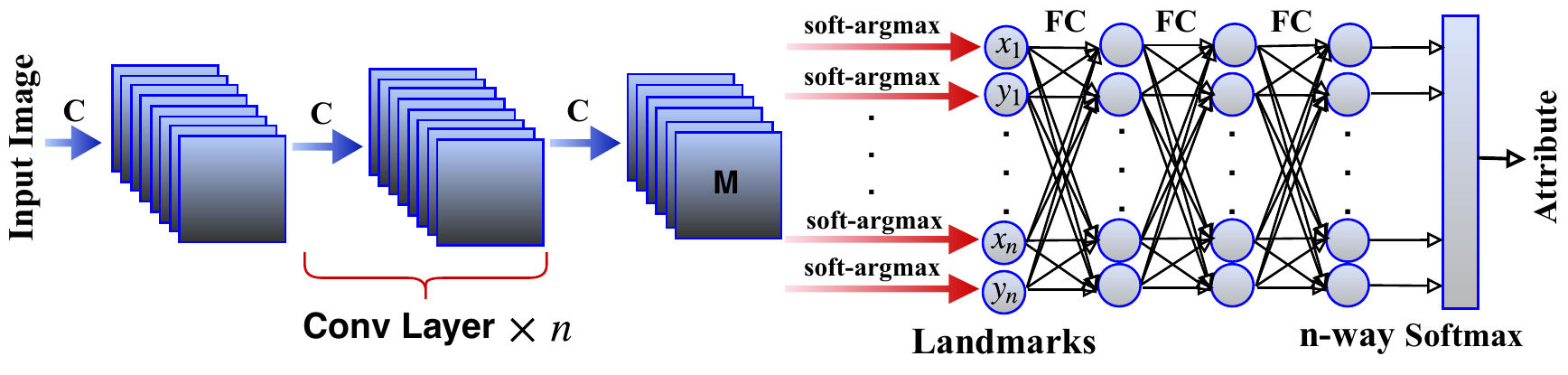}
\caption{\small Our basic implementation of the sequential multi-tasking architecture. The landmark localization model is composed a series of conv (C) layers (with no pooling) and a soft-argmax output layer to detect landmarks. $\times n$ indicates repeating conv layer $n$ times without parameter sharing. The detected landmarks are then fed to FC layers for attribute classification.
}
\label{fig:seq_multitask}
\end{center}
\vspace*{-1.em}
\end{figure}
\begin{figure}[t]
\begin{center}
\centering
{\includegraphics[width=0.06\textwidth]{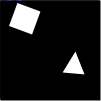}}
\fboxsep=0mm
\fboxrule=1pt
{\includegraphics[width=0.06\textwidth]
{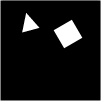}}
\fboxsep=0mm
\fboxrule=1pt
{\includegraphics[width=0.062\textwidth]
{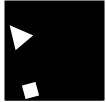}}
\fboxsep=0mm
\fboxrule=1pt
{\includegraphics[width=0.06\textwidth]
{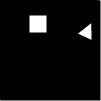}}
\fboxsep=0mm
\fboxrule=1pt
{\includegraphics[width=0.06\textwidth]
{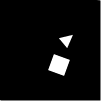}}\\
\vspace{.1em}
\fboxsep=0mm
\fboxrule=1pt
{\includegraphics[width=0.06\textwidth]{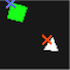}}
\fboxsep=0mm
\fboxrule=1pt
{\includegraphics[width=0.061\textwidth]
{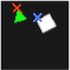}}
\fboxsep=0mm
\fboxrule=1pt
{\includegraphics[width=0.06\textwidth]
{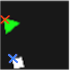}}
\fboxsep=0mm
\fboxrule=1pt
{\includegraphics[width=0.06\textwidth]
{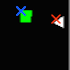}}
\fboxsep=0mm
\fboxrule=1pt
{\includegraphics[width=0.06\textwidth]
{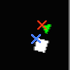}}\\
\vspace{.1em}
\fboxsep=0mm
\fboxrule=1pt
{\includegraphics[width=0.06\textwidth]
{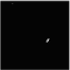}}
\fboxsep=0mm
\fboxrule=1pt
{\includegraphics[width=0.06\textwidth]
{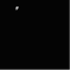}}
\fboxsep=0mm
\fboxrule=1pt
{\includegraphics[width=0.06\textwidth]
{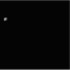}}
\fboxsep=0mm
\fboxrule=1pt
{\includegraphics[width=0.06\textwidth]
{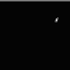}}
\fboxsep=0mm
\fboxrule=1pt
{\includegraphics[width=0.06\textwidth]
{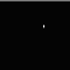}}\\
\vspace{.1em}
\fboxsep=0mm
\fboxrule=1pt
{\includegraphics[width=0.06\textwidth]
{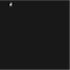}}
\fboxsep=0mm
\fboxrule=1pt
{\includegraphics[width=0.06\textwidth]
{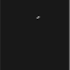}}
\fboxsep=0mm
\fboxrule=1pt
{\includegraphics[width=0.06\textwidth]
{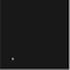}}
\fboxsep=0mm
\fboxrule=1pt
{\includegraphics[width=0.06\textwidth]
{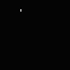}}
\fboxsep=0mm
\fboxrule=1pt
{\includegraphics[width=0.06\textwidth]
{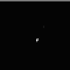}}
\end{center}
\vskip-10pt
    \caption{\small 
    Top row: Sample images from the Shapes dataset. Each $60 \times 60$ image contains one square and one triangle with randomly sampled location, size, and orientation.
    Second row: The two predicted landmarks and the object (in green) closest to the top-left corner classified by network.
    The \emph{third and fourth} show the first and second landmark feature maps, corresponding closely with the location of triangle and square. (Best viewed in color with zoom.)
}
\label{fig:shapes}
\end{figure}

To begin, we use a simple dataset to demonstrate our method's ability to learn consistent landmarks without direct supervision.
Images in our Shapes dataset (see Fig. \ref{fig:shapes} top row for examples) consist of a white triangle and a white square on black background, with randomly sampled size, location, and orientation.
The classification task is to identify which shape (triangle or square) is positioned closer to the upper-left corner of the image. We trained a model (as illustrated in Fig.~\ref{fig:seq_multitask}) with six convolutional layers using $7 \times 7$ kernels, followed by two convolutional layers with $1 \times 1$ kernels, then the soft-argmax layer for landmark localization. Predicted landmarks input to two fully connected (FC) layers of size $40$ and $2$, respectively. The model is trained with \emph{only} the cross-entropy cost on the class label \emph{without} labeled landmarks or the unsupervised ELT cost.

Figure~\ref{fig:shapes} shows the predictions of the trained model on a few samples from the dataset. In the second row, the green shape corresponds to the shape predicted to be the nearest to the upper-left corner, which was learned with $99\%$ accuracy.
The red and blue crosses correspond to the first soft-argmax and second soft-argmax landmark localizations, respectively.
We observe that the red cross is consistently placed adjacent to the triangle, while the blue cross is near the square. 
This experiment shows the sequential architecture proposed here properly guides the first part of the network to find meaningful landmarks on this dataset, based solely on the supervision of the related classification task.
\vskip-3pt

\subsection{Blocks dataset}
\label{sec:exp_blocks}
\begin{table}[h]
\caption{\small Error of different architectures on Blocks dataset. The error is reported in pixel space. An error of 1 indicates 1 pixel distance to the target landmark location. The first 4 rows show the results of Seq-MT architecture, as shown in Fig.~\ref{fig:seq_multitask}. The 5th and 6th rows show results of Comm-MT, depicted in Fig.~\ref{fig:common_multitask}. The last two rows show the results of Heatmap-MT, depicted in Fig. \ref{fig:paral_MT}. The results are averaged over five seeds.
}
\label{tab:blocks}
\vskip2pt
\centering
\resizebox{1\linewidth}{!}{
\begin{tabular}{c*{6}{C}}
\toprule
& \multicolumn{6}{c}{\textbf{Percentage of Images with Labeled Landmarks}} \\
\cmidrule{2-7}
\textbf{Model} & 1\% & 5\% & 10\% & 20\% & 50\% & 100\% \\
\midrule
Seq-MT (L) &  8.33 & 3.95 & 3.35 & 1.98 & 1.19 & 0.44\\
Seq-MT (L+A) &  8.02 & 3.45 & 3.20 & 1.67 & 1.05 & \textbf{0.38}\\
Seq-MT (L+ELT) &  6.42 & 1.94 & 1.37 & 1.16 & 0.85\\
\shortstack{Seq-MT (L+ELT+A)} & \textbf{6.25} & \textbf{1.70} & \textbf{1.26} & \textbf{1.07} & \textbf{0.74}\\
\midrule
Comm-MT (L) &  12.89 & 11.56 & 10.72 & 9.39 & 5.04 & 3.41\\
Comm-MT (L+A) &  12.28 & 11.19 & 10.36 & 9.01 & 4.21 & 2.97\\
\midrule
\shortstack{Heatmap-MT (L)} &  10.09 & 6.59 & 5.27 & 3.82 & 2.78 & 2.01\\
\shortstack{Heatmap-MT (L+A)} &  9.27 & 6.35 & 5.62 & 3.75 & 3.14 & 2.23\\
\bottomrule
\end{tabular}
}
\end{table}
\begin{figure}[h]
\begin{center}
\centering
\fboxsep=0mm
\fboxrule=1pt
\hspace{1.2em}\includegraphics[trim={0 0 16cm 0},clip,width=0.80\linewidth]{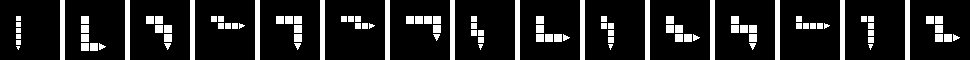}
\\
\hspace{1.2em}\includegraphics[trim={18.5cm 0 0 0},clip,width=0.70\linewidth]{images/blocks_classes.png} \\
\vspace{.4em}
\rotatebox{90}{\begin{minipage}[b]{26pt}{\begin{center}\tiny Seq-MT\\(L)\end{center}}\end{minipage}}\,{\includegraphics[width=0.35\textwidth]
{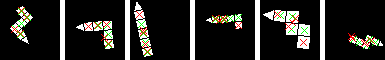}}\\
\fboxsep=0mm
\fboxrule=1pt
\rotatebox{90}{\begin{minipage}[b]{26pt}{\begin{center}\tiny Seq-MT\\(L+A)\end{center}}\end{minipage}}\,{\includegraphics[width=0.35\textwidth]
{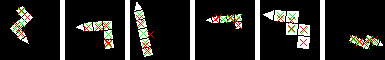}}\\
\fboxsep=0mm
\fboxrule=1pt 
\rotatebox{90}{\begin{minipage}[b]{26pt}{\begin{center}\tiny Seq-MT\\(L+ELT)\end{center}}\end{minipage}}\,{\includegraphics[width=0.35\textwidth]
{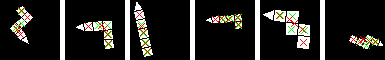}}\\
\fboxsep=0mm
\fboxrule=1pt
\rotatebox{90}{\begin{minipage}[b]{26pt}{\begin{center}\tiny Seq-MT\\(L+ELT+A)\end{center}}\end{minipage}}\,{\includegraphics[width=0.35\textwidth]
{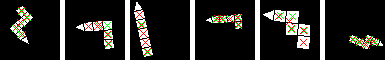}}
\end{center}
\vskip-6pt
    \caption{\small \emph{(top)} The fifteen classes of the Blocks dataset. Each class is composed of five squares and one triangle. To create each $60 \times 60$ image in the dataset, a random scale, translation, and rotation (up to $360$ degrees) is applied to one of the base classes. \emph{(bottom)} Sample landmark prediction on Blocks dataset using sequential multi-tasking models when only 5\% of data is labeled with landmarks. 
 Green and red cross show in order GT and predicted landmarks. (Best viewed in color with zoom.)
}
\label{fig:blocks}
\end{figure}
Our second toy dataset, Blocks, presents additional difficulty: each image depicts a figure composed of a sequence of five white squares with one white triangle at the head. See Fig.~\ref{fig:blocks}-top for all fifteen classes of Block dataset. 
We split the dataset into train, validation, and test sets, each having $3200$ images. 

Initially we trained the model with only cross-entropy on the class labels and evaluated the quality of the resulting landmark assignments. Ideally, the model would consistently assign each landmark to a particular block in the sequence from head (the triangle) to tail (the final square). However, in this more complex setting, the model did not predict landmarks consistently across examples. With the addition of the ELT cost, the model learns relatively consistent landmarks between examples \emph{from the same class}, but this consistency does not extend between different classes. Unlike the Shapes dataset---where there was a consistent, if indirect, mapping between landmarks and the classification task---the correspondence among the classification task and landmark identities is more tenuous in the Blocks dataset.
Hence, we introduce a labeled set of ground truth (GT) landmark locations and evaluate the landmark localization accuracy by having different percentages of the training set being labelled with landmarks.

Table~\ref{tab:blocks} compares the results using the sequential multi-tasking model in the following scenarios: 1) using only the landmarks (Seq-MT (L)), which is equivalent to training only the first part of the network, 2) using landmarks and attribute labels (Seq-MT (L+A)), which trains the whole network on class labels and the first part of the network on landmarks, 3) using landmarks and the ELT cost (Seq-MT (L+ELT)), which trains only the first part of the network, and 4) using three costs together (Seq-MT (L+ELT+A)).\footnote{The set of examples with labeled landmarks is class-balanced.} When using the ELT cost (scenarios 3 \& 4), we only apply it to images that do not provide GT landmarks to simulate semi-supervised learning 
\footnote{This is done to avoid unfair advantage of our model compared to other models on examples that provide landmarks. However, the ELT technique can be applied to any image, both with and without labeled landmarks.}.

As shown in Table \ref{tab:blocks}, the \emph{Seq-MT (L+A)} improves upon \emph{Seq-MT (L)}, indicating that class labels can be used to guide the landmark locations. By adding the ELT cost, we can improve the results considerably. With \emph{Seq-MT (L+ELT)} better performance is obtained compared to \emph{Seq-MT (L+A)} showing that the unsupervised learning technique can substantially enhance performance. However, the best results are obtained with all costs when using class labels, the ELT and landmark costs. See Fig. \ref{fig:blocks}-bottom for prediction samples when only $5\%$ of the data are labeled with landmarks.

\begin{figure}[t]
\centering
\includegraphics[width=\linewidth]{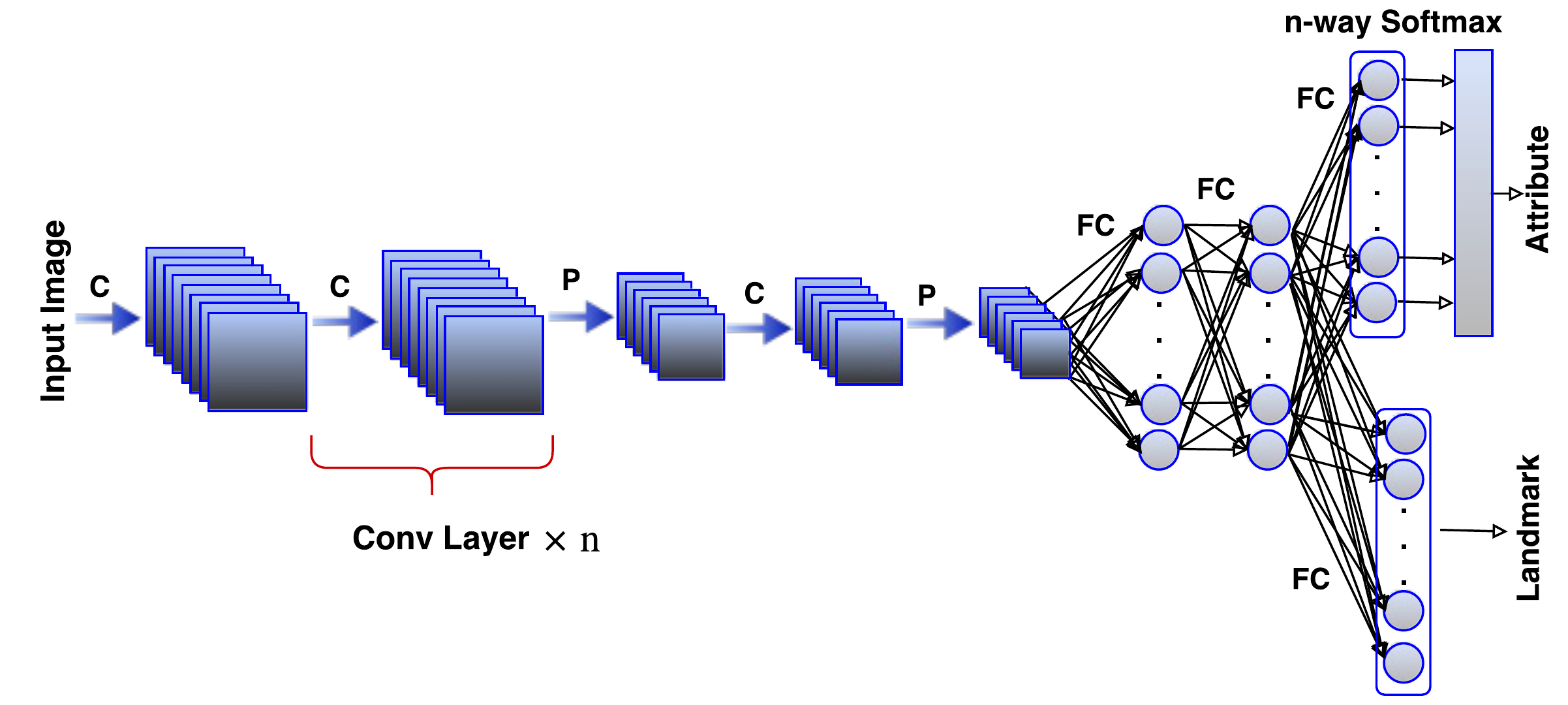}
\caption{\small Our implementation of the common multi-tasking (Comm-MT) architecture used in the literature \cite{zhang2014facialMTFL, zhang2016MAFL, zhang2014improving, devries2014multi}. 
The model takes an image and applies a series of conv (C) and pooling (P) layers which are then passed to few common (shared) FC layers. 
The last common FC layer is then connected to two branches (each for a task), one for the classification task and another for the landmark localization. 
}
\label{fig:common_multitask}
\end{figure}

\begin{figure}[t]
\begin{center}
\includegraphics[width=0.47\textwidth]{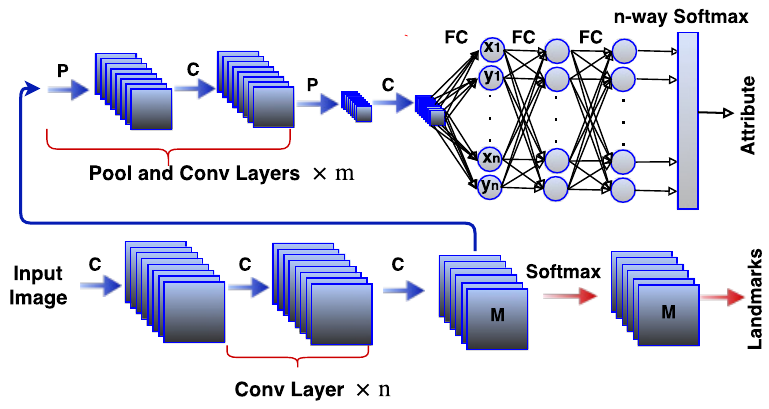}
\end{center}
\vskip-10pt
   \caption{\small Our implementation of multi-tasking architecture using heatmaps (Heatmap-MT). Landmarks are detected using conv (C) layers without sub-sampling, pooling, or FC layers. A softmax layer is used for landmark prediction in the output layer. Landmark heatmaps right before softmax layer are fed to a series of pool (P) and conv (C) layers which are then passed to FC layers. The last FC layer is fed to softmax for attribute classification.
}
\label{fig:paral_MT}
\end{figure}

Since our model can be considered as a multi-tasking network, we contrast it with other multi-tasking architectures in the literature. We compare with two architectures: 1) The ``common'' multi-tasking architecture (Comm-MT) \cite{zhang2014facialMTFL, zhang2016MAFL, zhang2014improving, devries2014multi} where sub-networks for each task share a common set of initial layers ending in a common fully-connected layer (see Fig.~\ref{fig:common_multitask}).\footnote{We tried other variants such as 1) a model that goes directly from the feature maps that have the same size as input image to the FC layer without any pooling layers and 2) a model that has more pooling layers and goes to a lower resolution before feeding the features to FC layers. Both models achieved worse results. Model 1 suffers from over-parameterization when going to FC layer. Model 2 suffers from loosing track of pooled features' locations since more pooling layers are used.}
We train two variants of this model, one with only landmarks (\emph{Comm-MT (L)}) and another with landmarks and class labels (\emph{Comm-MT (L+A)}) to see whether the class labels improve landmark localization. 2) Heat-map multi-tasking (Heatmap-MT), where -- to avoid pooling layers -- we follow the recent trend of maintaining the resolution of feature maps \cite{tompson2015efficient, hariharan2015hypercolumns, long2015fully, honari2016recombinator} and features detected for landmark localization do not pass through a FC layer. 
See Fig.~\ref{fig:paral_MT} for an illustration of this architecture. The heatmaps right before the softmax layer are taken as input to the classification model.
Note that this model doesn't have a landmark bottle-neck such as \emph{Seq-MT (L+A)}.

As shown in Table~\ref{tab:blocks}, 
the Comm-MT approach is performing much worse than our Seq-MT architecture for landmark estimation. A drawback of this architecture is its use of pooling layers which leads to sub-optimal results for landmark estimation. The model trained with extra class information performs better than the model trained only on landmarks. 
Heatmap-MT also performs worse than {Seq-MT}. This is likely due in part to Heatmap-MT using softmax log-likelihood training (which cannot be more accurate than the discretization grid), while Seq-MT uses soft-argmax training based on real number coordinates. Moreover, in {Heatmap-MT} the class label is mostly helping when using a low percentage of labeled data, but in {Seq-MT} it is helping for all percentages of labeled data. We believe this is due to creating a bottle-neck of landmarks before class label prediction, which causes the class labels to impact landmarks more directly through back-propagation.

\subsection{Hand pose estimation}
\label{sec:exp_hands}
\begin{figure}[ht]
\begin{tabular}{c}
{\includegraphics[width=0.45\textwidth]
{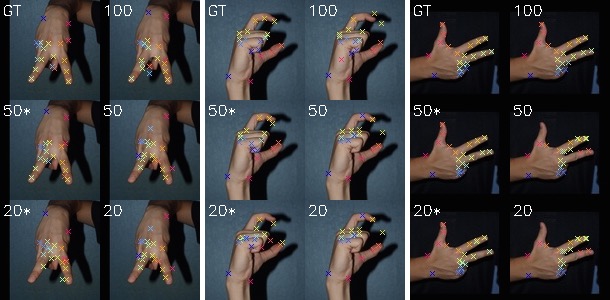}}
\end{tabular}
\caption{\small Examples of our model predictions on the test set of the HGR1 dataset \cite{kawulok2014PolishHands, nalepa2014PolishHands}. GT represents ground-trust annotations, while numbers 100, 50, and 20 indicate which percentage of the training set with labeled landmarks used for training. Results are computed with Seq-MT (L+ELT+A) model (denoted *) and Seq-MT (L). Best viewed in color with zoom.}
\label{fig:hands_examples}
\end{figure}

Our first experiment on real data is on images of hands captured with color sensors. Most common image datasets with landmarks on hands such as NYU \cite{tompson2014NYUhands} and ICVL \cite{tang2014ICVLhands} do not provide class labels. 
Also, most of the prior works in landmark estimation for hands are based on depth data \cite{yuan20183d, tompson2014NYUhands, tang2014ICVLhands, sridhar2013interactive, sun2015cascaded, sinha2016deephand} whereas estimating from color data is more challenging. 
We use the Polish hand dataset (HGR1) \cite{kawulok2014PolishHands, nalepa2014PolishHands}, which provides 898 RGB images with 25 landmarks and 27 gestures from Polish sign language captured with uncontrolled lightning and uncontrolled background from 12 subjects. We divide images by id (with no overlap in subjects between sets) into training set (ids 1 to 8), validation (ids 11 and 12), and test (ids 9 and 10). We end up with 573, 163, and 162 images for training, validation and test sets, respectively.
Accuracy of landmark detection on HGR1 dataset is measured by computing average RMSE metric in the image domain for every landmark and normalizing it by wrist width (the Euclidean distance between landmarks \#1 and \#25). We apply the ELT cost only on the images that do not have GT landmarks.
\begin{table}[t!]
\caption{\small Performance of architectures on HGR1 hands dataset. The error is Euclidean distance normalized by wrist width. Results are shown as percent; lower is better.}
\label{tab:hands}
\vskip2pt
\centering
\resizebox{.85\linewidth}{!}{
\begin{tabular}{c*{5}{C}}
\toprule
 & \multicolumn{5}{c}{\textbf{Percentage of Images with Labeled Landmarks}} \\ 
\cmidrule{2-6}
\textbf{Model} & 5\% & 10\% & 20\% & 50\% & 100\% \\
\midrule
Seq-MT (L) &	57.6 &	41.1 &	32.0 &	21.4 &	\textbf{15.8} \\
Seq-MT  (L+A) &	50.0 &	38.1 &	28.1 &	19.8 &	16.9 \\
Seq-MT  (L+ELT) &	43.7 &	31.5 &	25.1 &	\textbf{17.7} &\\
Seq-MT  (L+ELT+A) &\textbf{38.5} &	\textbf{30.3} &	\textbf{24.0} &	19.1 &\\
\midrule
Comm-MT (L) &	77.1 &	62.8 &	52.7 &	41.8 &	35.7\\
Comm-MT (L+A) &	53.4 &	39.3 &	35.5 &	26.9 &	24.1\\
\midrule
Heatmap-MT (L) &	66.5 &	51.9 &	42.4 &	30.9 &	25.5\\
Heatmap-MT (L+A) &	64.8 &	54.9 &	43.2 &	30.5 &	26.7\\
\bottomrule
\end{tabular}
}
\end{table}
Table~\ref{tab:hands} shows results for landmark localization on the HGR1 test set. All results are averaged over 5 seeds. We observe: 1) sequential multitasking improves results for most experiments compared to using only landmarks (Seq-MT(L)) or other multi-tasking approaches, 2) the ELT cost significantly improves results for all experiments, and 
3) \textit{Seq-MT (L+ELT+A)} compared to \textit{Seq-MT (L)} can achieve the same performance with only half provided landmark labels (see 5\%, 10\%, 20\%). 
We show examples of landmark prediction with different models in Fig.~\ref{fig:hands_examples}. 
ELT and attribute classification (A) losses significantly improve results with a smaller fraction of annotated landmarks.

\subsection{Multi-PIE dataset}
\label{sec:exp_MultiPIE}
\begin{figure}[h]
\begin{center}
\centering
\fboxsep=0mm
\fboxrule=1pt
\rotatebox{90}{\begin{minipage}[b]{130pt}{
\begin{center}\tiny 
\hspace{.8mm} Seq-MT \hspace{4.8mm} Seq-MT \hspace{5mm} Seq-MT \hspace{5mm} Seq-MT\\
\hspace{.25mm} (L) \hspace{5.5mm} (L+ELT+A)  \hspace{3.mm} (L+ELT)  \hspace{6.5mm} (L) \\
\hspace{1.mm} 5\% \hspace{8.mm} 5\%  \hspace{7.8mm} 20\%  \hspace{7.5mm} 100\%
\end{center}}\end{minipage}}\,
{\includegraphics[width=0.4\textwidth]
{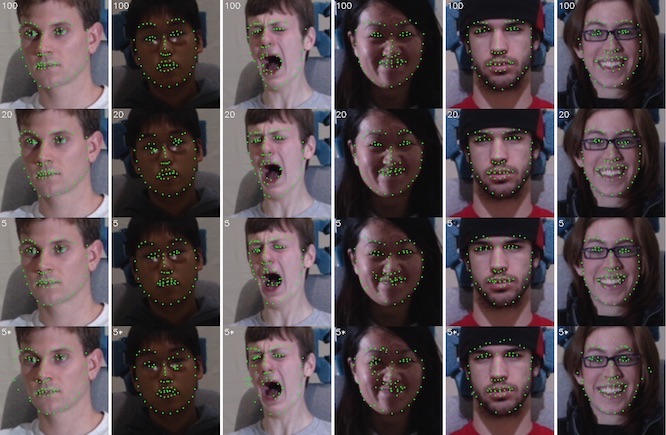}}
\end{center}
\vskip-6pt
\caption{\small Examples of our model predictions on Multi-PIE \cite{gross2010multi}.
On left you see the percentage of labelled data.
We observe close predictions between the top two rows indicating the effectiveness of the proposed ELT cost. Comparison between the last two rows shows the effectiveness of our method with only a small amount of labeled landmarks. Best viewed in color with zoom.
}
\label{fig:mpie_examples}
\end{figure}
We next evaluate our model on facial landmark datasets.
Similar to Hands, most common face datasets including Helen \cite{le2012Helen}, LFPW \cite{belhumeur2013LFPW}, AFW \cite{zhu2012AFW}, and 300W \cite{sagonas2013300}, only provide landmark locations and no classes. 
We start with Multi-PIE \cite{gross2010multi} since it provides, in addition to 68 landmarks, 6 emotions and 15 camera locations. 
We use
these as class labels to guide landmark prediction.\footnote{Our research does not involve face recognition, and emotion classes are used only to improve the precision of landmark localization.} 
We use images from 5 cameras (1 frontal, 2 with $\pm$15 degrees, and 2 with $\pm$30 degrees) and in each case a random illumination is selected.  The images are then divided into subsets by id\footnote{These ids are not personally identifiable information.}, with ids 1-150 in the training set, ids 151-200 in the valid set, and ids 201-337 in the test set. We end up with 1875, 579, and 1054 images in training, validation, and test sets. 

Due to using camera and emotion classes, our classification network has two branches, one for emotion and one for camera, with each branch receiving landmarks as inputs 
(see Supp. for architecture details). 
We compare our model with Comm-MT, Heatmap-MT architectures with and without class labels in Table \ref{tab:multiPIE}. 
Comparing models, we make the same observation as in Section \ref{sec:exp_hands} and the best performance is obtained when ELT and classification costs are used jointly, indicating both techniques are affective to get the least error. 
See some sample predictions in Fig. \ref{fig:mpie_examples}. 

\begin{table}[t]
\caption{\small Performance of different architectures on Multi-PIE dataset. The error is Euclidean distance normalized by eye-centers (as a percent; lower is better). We do not apply ELT cost on the examples that provide GT landmarks.}
\label{tab:multiPIE}
\vskip2pt
\centering
\resizebox{.85\linewidth}{!}{
\begin{tabular}{c*{5}{C}}
\toprule
 & \multicolumn{5}{c}{\textbf{Percentage of Images with Labeled Landmarks}} \\
\cmidrule{2-6}
\textbf{Model} & 5\% & 10\% & 20\% & 50\% & 100\% \\
\midrule
Seq-MT (L) & 7.98	&	7.02 &	6.28 &	5.50 &	\textbf{5.09}\\
Seq-MT (L+A) &  7.71 & 6.91 & 6.20 & 5.49 & 5.12 \\
Seq-MT (L+ELT) &  6.69 & 6.24 & 5.78 & 5.27 & \\
Seq-MT (L+ELT+A) & \textbf{6.57} & \textbf{6.16} & \textbf{5.73} & \textbf{5.23} &\\
\midrule
Comm-MT (L) & 9.22	& 7.93 & 7.02	& 6.27	& 5.71\\
Comm-MT (L+A) & 9.11	& 8.00	& 6.92	& 6.20	& 5.68 \\
\midrule
\shortstack{Heatmap-MT (L)} & 10.83  & 9.18 & 8.13 & 7.00 & 6.63 \\
\shortstack{Heatmap-MT (L+A)} & 11.03  & 9.03 & 8.15 & 7.11 & 6.65 \\
\bottomrule
\end{tabular}
}
\end{table}
 
\subsection{300W dataset}
\label{sec:exp_300W}
\begin{table*}[ht]
\caption{\small Comparison of recent models on their training conditions. RAR and Lv et. al \cite{Lv_2017_CVPR} initialize their models by pre-trained parameters. 
TCDCN uses 20,000 extra labelled data. Finally, RAR adds manual samples by occluding images with sunglasses, medical masks, phones, etc to make them robust to occlusion. Similar to RCN, Seq-MT and RCN$\stackanchor{+}{}$ both have an explicit validation set for HP selection and therefore use a smaller training set. Neither use any extra data, either through pre-trained models or explicit external data.
}
\label{tab:comp_models}
\vskip-4pt
\centering
\resizebox{.9\linewidth}{!}{
\begin{tabular}{cccccccc}
\toprule
 & \multicolumn{7}{c}{\textbf{Model}} \\
\cmidrule{2-8}
\textbf{Feature} & RAR \cite{xiao2016robust} & Lv et. al \cite{Lv_2017_CVPR} & TCDCN \cite{zhang2015learning} & CFSS \cite{zhu2015face} & RCN \cite{honari2016recombinator} & Seq-MT / RCN$\stackanchor{+}{}$ & RCN$\stackanchor{+}{}$ (L+ELT) (all-train) \\
\midrule
Hyper-parameter selection dataset & {\color{red}Test-set} & {\color{red}Test-set} & {\color{red}Test-set} & {\color{red}Test-set} & {\color{green}Valid-set} & {\color{green}Valid-set} & {\color{red}Test-set}\\
Training on entire training set? &  {\color{red}Yes} & {\color{red}Yes} & {\color{red}Yes} & {\color{red}Yes} & {\color{green}No} & {\color{green}No} & {\color{red}Yes}\\
Uses extra dataset? &  {\color{red}Yes} & {\color{red}Yes} & {\color{red}Yes} & {\color{green}No} & {\color{green}No} & {\color{green}No} & {\color{green}No} \\
Manually augmenting the training set? & {\color{red}Yes} & {\color{green}No} & {\color{green}No} & {\color{green}No} & {\color{green}No} & {\color{green}No} & {\color{green}No}\\
FPS (GPU) & 4 & 83 & 667 & - & 545 & 487 / 545 & 545\\
\end{tabular}
}
\end{table*}
In order to evaluate our architecture on natural images in the wild we use 300W \cite{sagonas2013300} dataset. This dataset provides 68 landmarks and is composed of 3,148 (337 AFW, 2,000 Helen, and 811 LFPW) and 689 (135 IBUG, 224 LFPW, and 330 Helen) images in the training and test sets, respectively. Similar to RCN \cite{honari2016recombinator}, we split the training set into 90\% (2,834 images) train-set and 10\% (314 images) valid-set. Since this dataset does not provide any class label, we can evaluate our model in L and L+ELT cases. 

In Table~\ref{tab:face_wild}-left we compare Seq-MT with other models in the literature. Seq-MT model is outperforming many models including CDM, DRMF, RCPR, CFAN, ESR, SDM, ERT, LBF and CFSS, and is only doing worse than few recent models with complicated architectures, e.g., RCN \cite{honari2016recombinator} with multiple branches, RAR \cite{xiao2016robust} with multiple refinement procedure and Lv et.~al.\  \cite{Lv_2017_CVPR} with multiple steps. 
Note that the originality of Seq-MT is not in the specific architecture used for the first part of the network that localizes landmarks, but rather in its multi-tasking architecture (specifically in its usage of the class labels to enhance landmark localization) and also leveraging ELT cost. The landmark localization part of Seq-MT can be replaced with more complex models. To verify this, we use the RCN model \cite{honari2016recombinator}, with publicly available code, and replace the original softmax layer with a soft-argmax layer in order to apply the ELT cost. We refer to this model as RCN$\stackanchor{+}{}$ and it is trained with these hyperparameters: $\beta = 1.0$, $\alpha = 0.5$, $\gamma = 0$, $\lambda = 1.0$. The result is shown as RCN$\stackanchor{+}{}$(L) when using only landmark cost and RCN$\stackanchor{+}{}$(L+ELT) when using landmark plus ELT cost. On 300W dataset we apply the ELT cost to samples with or without labelled landmarks to observer how much improvement can be obtained when used on all data. We can further reduce RCN error from 5.54 to 5.1 by applying the ELT cost and soft-argmax. This is a new state of the art without any data-augmentation. Also we evaluate accuracy of RCN$\stackanchor{+}{}$(L+ELT) trained without validation set and with early stopping on test set and achieve error of 4.9 - the overall state-of-the-art on this dataset. 

\begin{figure}[ht]
\begin{center}
\centering
\fboxsep=0mm
\fboxrule=1pt
\rotatebox{90}{\begin{minipage}[b]{130pt}{
\begin{center}\tiny 
\hspace{.8mm} RCN $\stackanchor{+}{}$ \hspace{4.8mm} RCN $\stackanchor{+}{}$ \hspace{5mm} Seq-MT \hspace{5mm} Seq-MT\\
\hspace{-2.8mm} (L+ELT) \hspace{6mm} (L)  \hspace{6.5mm} (L+ELT)  \hspace{6.5mm} (L) \\
\end{center}}\end{minipage}}\,
{\includegraphics[width=0.4\textwidth]
{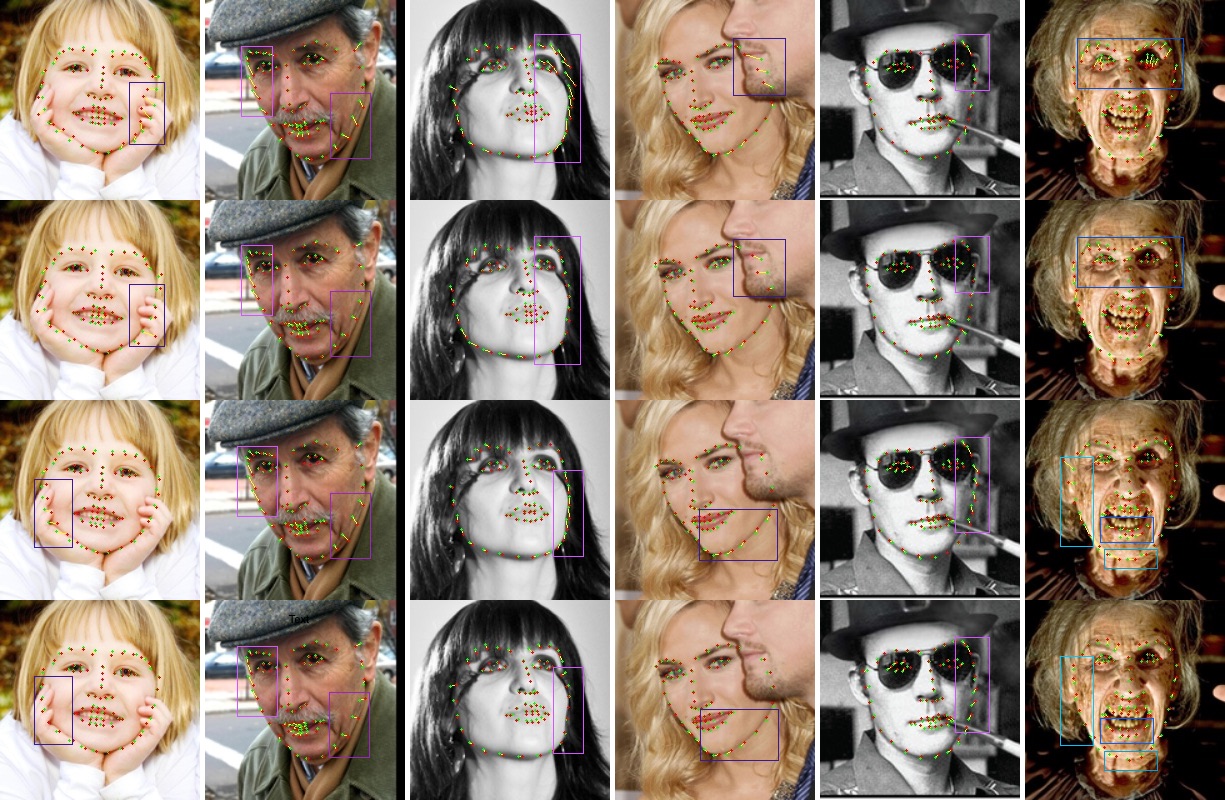}}
\end{center}
\vskip-6pt
\caption{\small Landmark localization samples on 300W \cite{sagonas2013300} test-set. 
The green and red dots show GT and model predictions, respectively. The yellow lines show the error. These examples illustrate the improved accuracy obtained by using the ELT cost. The rectangles show the regions that landmarks are mostly improved.
}
\label{fig:300W_examples}
\end{figure}

In Table~\ref{tab:300W_percent} we compare Seq-MT with Heatmap-MT and Common-MT 
on different percentage of labelled landmarks. We also demonstrate the improvement that can be obtained by using RCN$\stackanchor{+}{}$. Note that the ELT cost improves the results when applied to two different landmark localization architectures (Seq-MT, RCN). Moreover, it considerably improves the results on IBUG test-set that contains more difficult examples than the training set. Figure \ref{fig:300W_examples} shows the improvement obtained by using ELT cost on some test set samples.

\begin{table}[t]
\caption{\small Comparison with other SOTA models (as a percent; lower is better). \emph{(left)} Performance of different architectures on 300W test-set using 100\% labeled landmarks. The error is Euclidean distance normalized by ocular distance. 
\emph{(right-top)} Comparison with four other multi-tasking approaches and RCN. For these comparisons, we have implemented the specific architectures proposed in those papers. Error is as in Sections \ref{sec:exp_hands} and \ref{sec:exp_MultiPIE}.
\emph{(right-bottom)} Comparison of different architectures on AFLW test set. The error is Euclidean distance normalized by face size.}
\label{tab:face_wild}
\vskip-2pt
\hskip-40pt
\centering
\resizebox{.5\linewidth}{!}{
\begin{tabular}{cccc}
\multicolumn{4}{c}{\fontsize{13}{15} \textbf{300W Dataset}} \\
\toprule
\textbf{Model} & \textbf{Common} & \textbf{IBUG} & \textbf{Fullset} \\
\midrule
CDM \cite{yu2013} & 10.10 & 19.54 & 11.94 \\
DRMF \cite{asthana2013robust} & 6.65 & 19.79 & 9.22\\
RCPR \cite{burgos2013} & 6.18 & 17.26 & 8.35\\
CFAN \cite{zhang2014coarse} & 5.50 & 16.78 & 7.69\\
ESR \cite{cao2014face} & 5.28 & 17.00 & 7.58\\
SDM \cite{xiong2013} & 5.57 & 15.40 & 7.50 \\
ERT \cite{cao2014face} & & & 6.40 \\
LBF \cite{ren2014face} & 4.95 & 11.98 & 6.32\\
CFSS \cite{zhu2015face} & 4.73 & 9.98 & 5.76 \\
TCDCN* \cite{zhang2015learning} & 4.80 & 8.60 & 5.54\\
RCN \cite{honari2016recombinator} & 4.70 & 9.00 & 5.54 \\
\shortstack{RCN +\textbackslash \\ denoising} \cite{honari2016recombinator} & 4.67 & 8.44 & 5.41 \\
RAR \cite{xiao2016robust} & \textbf{4.12} & 8.35 & 4.94\\
Lv et. al \cite{Lv_2017_CVPR} & 4.36 & \textbf{7.56} & 4.99 \\
\midrule
Heatmap-MT (L) & 6.18 & 13.56 & 7.62\\
Comm-MT (L) & 5.68 & 11.04 & 6.73\\
Seq-MT (L) & 4.93 & 10.24 & 5.95\\
Seq-MT (L+ELT) & 4.84 & 9.53 &	5.74\\
\midrule
RCN$\stackanchor{+}{}$ (L) & 4.47 & 8.47 & 5.26\\
RCN$\stackanchor{+}{}$ (L+ELT) & 4.34 & 8.20 & 5.10\\
\shortstack{RCN$\stackanchor{+}{}$ (L+ELT)\\ (all-train)} & 4.20 & 7.78 & \textbf{4.90}\\
\end{tabular}
}
\resizebox{.35\linewidth}{!}{
\centering
\begin{minipage}{0.3\textwidth}
\begin{tabular}{c|cc|c}
Model & \multicolumn{2}{c|}{\textbf{Multi-PIE}} & \textbf{{HGR1}}  \\ \hline
Percent Labelled  & 5\%  & 100\% & 100\% \\ \hline
MT-DCNN \cite{zhang2014improving}(L+A)   & 11.13 & 7.60  & 20.87   \\
TCDCN \cite{zhang2014facialMTFL}(L+A)   & 18.46 & 10.59 & 25.85    \\ 
TCDCN-2 \cite{zhang2016MAFL}(L+A)   & 10.75 & 5.83  & 18.81 \\ 
MT-Conv \cite{devries2014multi}(L+A) & 9.99 & 8.08   & 19.20 \\ \hline
RCN \cite{honari2016recombinator} (L)   & 7.53 & 5.78   & 13.65 \\ \hline
RCN+ (L)  & 6.89 & 5.04   & 11.02 \\ 
RCN+ (L+A)  & \textbf{6.82} & \textbf{4.97} & \textbf{10.88} \\
\end{tabular}
\vskip20pt
\hskip30pt
\begin{tabular}{cccc}
\multicolumn{4}{c}{\fontsize{13}{15} \textbf{AFLW Dataset}} \\
\toprule
\textbf{Model} & \multicolumn{3}{c}{\textbf{Labeled Images}} \\
& 1\% & 5\% & 100\% \\ 
\midrule
CDM \cite{yu2013}  & \_ & \_ & 5.43 \\
ERT \cite{cao2014face} & \_ & \_ & 4.35 \\
LBF \cite{ren2014face} & \_ & \_ & 4.25 \\
SDM \cite{xiong2013} & \_ & \_ & 4.05 \\
CFSS \cite{zhu2015face} & \_ & \_ & 3.92 \\
RCPR \cite{burgos2013} & \_ & \_ & 3.73 \\
CCL \cite{Zhu_2016_CVPR} & \_ & \_ & 2.72 \\
Lv et. al \cite{Lv_2017_CVPR}  & \_ & \_ & 2.17 \\
\midrule
RCN$\stackanchor{+}{}$ (L) & 2.88 &  2.17& 1.61\\
RCN$\stackanchor{+}{}$ (L+A) & 2.52& 2.08& 1.60\\
\shortstack{RCN$\stackanchor{+}{}$ \\ (L+ELT+A)} &  2.46&  \textbf{2.03}& \textbf{1.59}\\
\end{tabular}
\end{minipage}
}
\end{table}
\vskip-5pt
\begin{table}[t]
\caption{\small Performance of different architectures on 300W test-set. The error is Euclidean distance normalized by ocular distance (eye-centers). Error is shown as a percent; lower is better.}
\label{tab:300W_percent}
\vskip-2pt
\centering
\resizebox{.85\linewidth}{!}{
\begin{tabular}{cc*{5}{C}}
\toprule
 & & \multicolumn{5}{c}{\textbf{Percentage of Images with Labeled Landmarks}} \\
\cmidrule{3-7}
& \textbf{Model} & 5\% & 10\% & 20\% & 50\% & 100\% \\
\bottomrule
\hline
\\[-2ex]
\multirow{6}{*}{\textbf{Fullset}}
& Heatmap-MT (L) & 13.47 & 11.68 & 9.85 & 8.18 & 7.62\\
& Comm-MT (L) & 16.73 & 9.66 & 8.61 & 7.39 & 6.73\\
& Seq-MT (L) & 9.82 & 8.30 &	7.26 &	6.28 &	5.95\\  
& Seq-MT (L+ELT) & 8.23 & 7.28 & 6.62 & 6.10 & 5.74\\
\cmidrule{2-7}
& RCN$\stackanchor{+}{}$ (L) & 7.26 & 6.48 &	5.91 &	5.52 &	5.26\\  
& RCN$\stackanchor{+}{}$ (L+ELT) & \textbf{7.22} & \textbf{6.32} & \textbf{5.88} & \textbf{5.45} & \textbf{5.10}\\
\bottomrule
\hline
\\[-2ex]
\multirow{6}{*}{\textbf{IBUG}}
& Heatmap-MT (L) & 26.36 & 22.77 & 18.46 & 14.94 & 13.56\\
& Comm-MT (L) & 28.64 & 16.17 & 14.56 & 12.16 & 11.04\\
& Seq-MT (L) & 18.74 & 16.21 &	13.41 &	11.20 &	10.24 \\
& Seq-MT (L+ELT) & 14.68 & 12.73 & 11.39 & 10.37 & 9.53\\
\cmidrule{2-7}
& RCN$\stackanchor{+}{}$ (L) & 15.36 & 12.74 &	11.82 &	10.12 &	8.47 \\
& RCN$\stackanchor{+}{}$ (L+ELT) & \textbf{12.54} & \textbf{10.35} & \textbf{9.56} & \textbf{8.67} & \textbf{8.20}\\
\end{tabular}
}
\end{table}

\subsection{AFLW dataset}
\label{sec:exp_AFLW}

AFLW \cite{AFLW_2011} contains images of 24,386 faces with 19 fiducial landmarks and 3D real-valued head pose information. We use pose as auxiliary task.
We split dataset into training,  testing sets, with 20,000 and 4,386 images, respectively. Furthermore, we allocate 2,000 images from training set for validation set. We use the same splits as in previous work \cite{ren2014face}, \cite{Lv_2017_CVPR} for direct comparison. We normalize RMSE by face size as in \cite{Lv_2017_CVPR}. 
We evaluate our method on RCN$\stackanchor{+}{}$ trained with ELT cost and head pose regression cost and obtain a new state of the art of $1.59$ with 27\% relative improvement. See comparison with other models in Table~\ref{tab:face_wild}-righ-bottom. We also evaluate our method with only 180 (1\%) or 900 (5\%) images of labeled landmarks. Under these settings we get significant improvement  with semi-supervised learning. With only 5\% of labeled data our method outperforms the previous state of the art methods. 
\subsubsection{Comparison with other techniques}
\label{sec:exp_comparison}
\vskip-5pt
In Table~\ref{tab:face_wild} we compare with recent models proposed for landmark localization and in Table~\ref{tab:comp_models} we evaluate their training conditions. RAR, TCDCN, Lv et.~al., and CFSS do not use an explicit validation set. This makes comparison with these models more difficult for two reasons: 1) These models do hyper-parameter (HP) selection on the test set, which makes them overfit on the test set; and 2) Their effective training size is bigger. When we use the entire training set (row \emph{RCN$\stackanchor{+}{}$ (L+ELT) (all-train)} in Table~\ref{tab:face_wild}-left we report new SOTA on 300W dataset. The first three models use extra datasets, either through pre-trained models (RAR, Lv et.~al.) or additional labeled data (TCDCN), while we do not leverage any extra data. 
Finally, our method is 136 and 6.5 times faster than RAR and Lv~et.~al methods.
\vskip-30pt
\section{Conclusion}
We presented a new architecture and training procedure for semi-supervised landmark localization. Our contributions are twofold;
We first proposed an unsupervised technique that leverages equivariant landmark transformation without requiring labeled landmarks. 
In addition we developed an architecture to improve landmark estimation using auxiliary attributes such as class labels by backpropagating errors through the landmark localization components of the model.
Experiments show that these achieve high accuracy with far fewer labeled landmark training data in tasks of landmark location for hands and faces. We achieve new state of the art performance on public benchmark datasets for fiducial points in the wild, 300W and AFLW. 
\vspace*{-0.2cm}
\subsubsection*{Acknowledgment}
\vspace*{-0.1cm}
We would like to thank Compute Canada and Calcul Quebec for providing computational resources. This work was partially funded by NVIDIA’s NVAIL program.

{\small
\bibliographystyle{ieee}
\bibliography{egbib}
}

\clearpage
\renewcommand{\thesection}{S.\arabic{section}}
\renewcommand{\thesubsection}{\thesection.\arabic{subsection}}

\newcommand{\beginsupplementary}{%
  \setcounter{table}{0}
  \renewcommand{\thetable}{S\arabic{table}}%
  \setcounter{figure}{0}
  \renewcommand{\thefigure}{S\arabic{figure}}%
}

\beginsupplementary
\twocolumn[{%
 \centering
\textbf{\Large Supplementary Information for \\Improving Landmark Localization with Semi-Supervised Learning\\}
\vspace{3. em}
}]

\subsection{Comparison on MTFL dataset}
In table \ref{Compare_MTFL} we compare with other models on MTFL \cite{zhang2014facialMTFL} dataset which provides 5 landmarks on facial images: eye-centers, nose tip, mouth corners. We follow the same protocol as \cite{honari2016recombinator} for comparison, where we use train and valid sets of 9,000 and 1,000 images, respectively. We test our model on AFLW and AFW subsets, with 29,995 and 337 images, that were re-annotated with 5 landmarks. For the $L+A$ case we use the head-pose which is categorized into one of the five cases: right profile, right, frontal, left, left profile. Other attribute labels, e.g. gender and wearing glasses, cannot be determined from such few landmarks and therefore are not useful in our proposed semi-supervised learning of landmarks.
\begin{table}[htb!]
\centering
\caption{Results on MTFL test sets for 100\% labelled data}
\label{Compare_MTFL}\vspace{-2mm}
\resizebox{.99\linewidth}{!}{
\begin{tabular}{c|ccccc|cc}
 & \multicolumn{5}{c}{Model} & \multicolumn{2}{|c}{Our} \\
 & ESR& RCPR & SDM & TCDCN & RCN & RCN+(L) & RCN+(L+A) \\ \hline
AFLW & 12.4 & 11.6 & 8.5 & 8.0 & 5.6 & 5.22 & \textbf{5.02} \\
AFW & 10.4 & 9.3 & 8.8 & 8.2 & 5.36 & 5.13 & \textbf{5.08} \\
\end{tabular}
}
\end{table}
\subsection{Selecting auxiliary labels for semi-supervised learning}
The impact of an attribute on the landmark in sequential training depends on the amount of informational overlap between the attribute and the landmarks. We suggest to measure the normalized mutual information adjusted to randomness (Adjusted Mutual Information (AMI)), as a selection heuristic, prior to applying our method. AMI ranges from 0 to 1 and indicates the fraction of statistical overlap. 
We compute for each attribute its AMI with all landmark coordinates. 

On Multi-PIE we got AMI(x;y) = 0.045, indicating a low mutual information between coordinates x and y. We therefore compute AMI for attribute (A) and every landmark (as x,y pair) by discretizing every variable uniformly under assumption of coordinate independence: AMI(A;x,y) = AMI(A;x) + AMI(A;y). Every variable is uniformly discretized to have 20 levels at most. Finally we measure averaged mutual information between an attribute and the set of landmarks as 
$$\frac{1}{N \times L} \sum_{n \in N} \sum_{x_n \in X_n, y_n \in Y_n} AMI(A_n;x_n) + AMI(A_n;y_n)$$
where $N$ and $L$ indicate the number of samples and landmarks, $X_n$ and $Y_n$ indicate the set of $x$ and $y$ landmark coordinates per sample $n$. 
In Table~\ref{mutual_info} we observe that hand gesture labels and head pose regression are among the most effective attributes for our method. There is little mutual information between wearing glasses and landmarks, indicating lack of usefulness of this attribute for our semi-supervised setting. 
\begin{table}[htb!]
\centering
\caption{Mutual Information between all landmarks and each attribute}
\label{mutual_info}\vspace{-3mm}
\resizebox{1.\linewidth}{!}{
\begin{tabular}{c|c|c|c|c|c}
Dataset  & \multicolumn{1}{c|}{}       &  \multicolumn{3}{c|}{MultiPIE} & HGR1 \\ \hline
\shortstack{Attribute} & Random & Emotion & Camera & Identity & Gesture Label \\ \hline
AMI, mean & .000 & .098 & .229 & .049 & .559\\
AMI, max  & .006 & .229 & .493 & .088 & .669\\ 
\end{tabular} 
}
\vfill
\vspace{4pt}
\resizebox{1.\linewidth}{!}{
\begin{tabular}{c|c|c|c|c}
Dataset  & \multicolumn{1}{c|}{}   & AFLW & \multicolumn{2}{c}{MTFL} \\ \hline
\shortstack{Attribute} & Random & Pose Regression & Glasses & Pose Classification  \\ \hline
AMI, mean & .000 & .536 & .002 & .069\\
AMI, max  & .006 & .576 & .003 & .222\\ 
\end{tabular}
}
\end{table}

The attributes that are mostly useful yield a high accuracy, or low error, if we just train a neural network that takes only ground truth landmarks as input and predicts the attribute. This indicates that by relying only on landmarks we can get high accuracy for those attributes. In Table \ref{GT_predict} we compare the attribute prediction accuracy from the proposed Seq-MT model with a case when we do such prediction from GT landmarks. Prediction from GT landmarks always outperforms the one of Seq-MT. This indicates that in our semi-supervised setting, where we have few labelled landmarks, by improving the predicted locations of landmarks, both attribute and landmarks error would reduce. 
\begin{table}[htb!]
\centering
\caption{\small Attribute classification accuracy (MultiPIE, HGR1)---higher is better---or prediction error (AFLW)---lower is better---from GT \& estimated landmarks.}
\label{GT_predict}\vspace{-3mm}
\resizebox{1\linewidth}{!}{
\begin{tabular}{c|c|c|c|c}
       & \multicolumn{2}{c|}{MultiPIE} & HGR1 & AFLW\\ \hline
Attribute      & Camera & Emotion & Label  & Pose Error\\ \hline
\shortstack{From GT Landmarks}  & 99.54 $\textcolor{green}{\uparrow}$ & 88.21 $\textcolor{green}{\uparrow}$ & 91.7 $\textcolor{green}{\uparrow}$ & 4.98 $\textcolor{green}{\downarrow}$ \\ \hline
\shortstack{Best Seq-MT Attr. Predict.}  & 98.96$\quad$ & 86.48$\quad$ & 79.1$\quad$ & 5.10$\quad$ \\
\end{tabular}
}
\end{table}
\vspace{4pt}
\subsection{Comparison of softmax and soft-argmax}
Heatmap-MT(L) and Seq-MT(L) have the same architectures but use different loss functions (softmax vs. soft-argmax). RCN(L) and RCN+(L) also only differ in their loss function.
When comparing these models in Tables \ref{tab:blocks}, \ref{tab:hands}, \ref{tab:multiPIE}, \ref{tab:face_wild}, and \ref{tab:300W_percent} soft-argmax outperforms soft-max. To further examine these two losses we replace soft-max with soft-argmax in Heatmap-MT and show the results in Table \ref{heatmap_softarg}. Comparing the results in Table \ref{heatmap_softarg} with Tables \ref{tab:hands} and \ref{tab:multiPIE}, we observe improved performance of landmark localization using soft-armgax. In soft-max the model cannot be more accurate than the number of elements in the grid, since soft-max does a classification over the pixels. However, in soft-argmax the model can regress to any real number and hence can get more accurate results. We believe this is the reason behind its better performance.
\begin{table}[htb!]
\centering
\caption{Results on Heatmap-MT (L+A) comparing soft-max with soft-argmax.}
\label{heatmap_softarg}\vspace{-2mm}
\resizebox{0.9\linewidth}{!}{
\begin{tabular}{c|c|c|c|c|c|c}
Dataset    & & 5\%   & 10\%  & 20\%  &  50\%  & 100\%  \\  \hline
\multirow{ 2}{*}{Multi-PIE} & softmax & 11.03  & 9.03 & 8.15 & 7.11 & 6.65 \\ 
& soft-argmax  & \textbf{8.00}  & \textbf{7.06}  & \textbf{6.29}  &  \textbf{5.49}  & \textbf{5.14}   \\ \hline
\multirow{ 2}{*}{HGR1}&  softmax &	64.8 &	54.9 &	43.2 &	30.5 &	26.7 \\
& soft-argmax & \textbf{56.88} & \textbf{42.79} & \textbf{33.07} & \textbf{22.5}  & \textbf{18.8} \\
\end{tabular}
}
\end{table}
\vspace{-5pt}
\subsection{Supplementary results on Multi-PIE dataset}
Although the focus of this paper is on improving landmark localization, in order to observe the impact of each multi-tasking approach on the attribute classification accuracy, we report the classification results on emotion in Table \ref{tab:multiPIE_emotion} and on camera in Table \ref{tab:multiPIE_camera}. Results show that the classification accuracy improves by providing more labeled landmarks, despite having the number of \emph{(image, class label)} pairs unchanged. 
It indicates that improving landmark localization can directly impact the classification accuracy.
Landmarks are especially more helpful in emotion classification. On camera classification, the improvement is small and all models are getting high accuracy. Another observation is that Heatmap-MT performs better on classification tasks compared to the other two multi-tasking approaches. We believe this is due to passing more high-level features from the image to the attribute classification network compared to Seq-MT. However, this model is performing worse than Seq-MT on landmark localization. The Seq-MT model benefits from the landmark bottleneck to improve its landmark localization accuracy. In Tables \ref{tab:multiPIE_emotion} and \ref{tab:multiPIE_camera} by adding the ELT cost the classification accuracy improves (in addition to landmarks) indicating the improved performance in landmark localization can enhance classification performance. 

\begin{figure*}[ht]
\begin{center}
\begin{tabular}{c}
\rotatebox{90}{\begin{minipage}[b]{205pt}{
\begin{center}\tiny 
\hspace{.8mm} Seq-MT \hspace{6mm}\hspace{4.8mm} Seq-MT \hspace{8mm}\hspace{5mm} Seq-MT \hspace{6mm}\hspace{5mm} Seq-MT\\
\hspace{.25mm} (L) \hspace{6mm}\hspace{5.5mm} (L+ELT+A)  \hspace{8mm}\hspace{3.mm} (L+ELT)  \hspace{6mm}\hspace{6.5mm} (L) \\
\hspace{1.mm} 5\% \hspace{6mm}\hspace{8.mm} 5\%  \hspace{8mm}\hspace{7.8mm} 20\%  \hspace{6mm}\hspace{7.5mm} 100\%
\end{center}}\end{minipage}}\,
\includegraphics[width=0.95\linewidth]{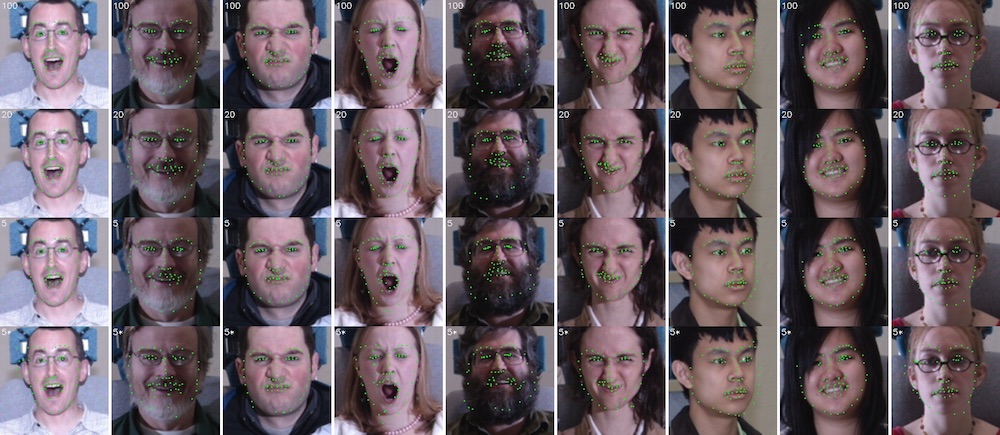}
\vspace{1.5em}
\end{tabular}
   \caption{Extra examples of our model predictions on Multi-PIE \cite{gross2010multi} test set. 
We observe close predictions by 1) and 2) indicating the effectiveness of our proposed ELT cost even with only a small amount of labeled landmarks.  Comparison between 3) and 4) shows the improvement obtained with both the ELT loss and the sequential multitasking architecture when using a small percentage of labeled landmarks. Note that the model trained with ELT loss preserves better the joint distribution over the landmarks even with a small number of labeled landmarks. The last two examples show examples with high errors. Best viewed in color with zoom.}
\label{fig:MultiPIE_examples_2}
\end{center}
\end{figure*}

Figure \ref{fig:MultiPIE_examples_2} provides further localization examples on Multi-PIE dataset. 

\subsection{Supplementary results on hands dataset}
In Table \ref{tab:hands_class} we show classification accuracy obtained using different multi-tasking techniques. Similar to the Multi-PIE dataset, we observe increased accuracy by providing more labeled landmarks, showing the classification would benefit directly from landmarks. Also similar to Multi-PIE, we observe better classification accuracy with Heatmap-MT. Comparing Seq-MT models, we observe improved classification accuracy by using the ELT cost. It demonstrates the impact of this component on both landmark localization and classification accuracy.

\begin{figure*}[ht]
\begin{center}
\begin{tabular}{c}
\includegraphics[width=0.95\linewidth]{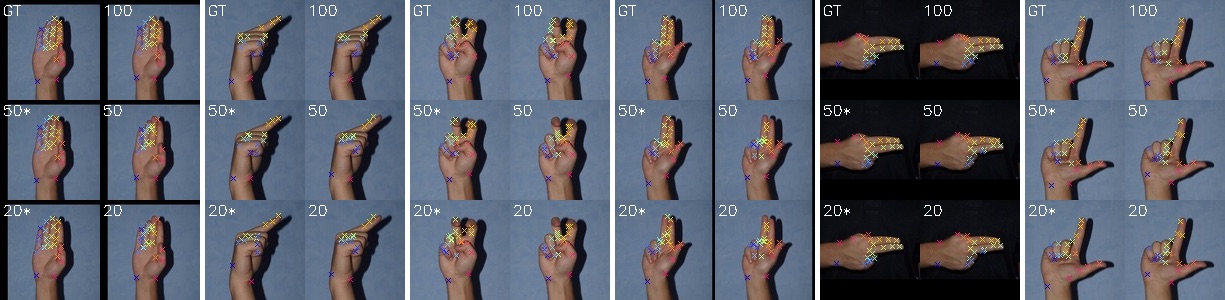}
\vspace{.5em}
\\
\includegraphics[width=0.95\linewidth]{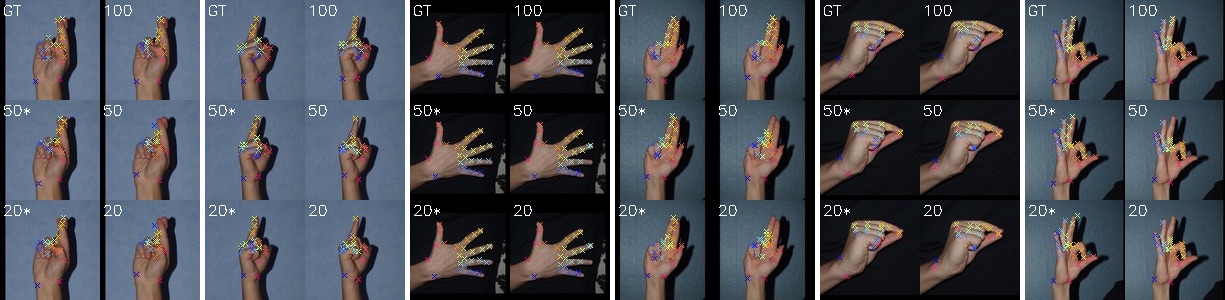}
\vspace{.5em}
\\
\includegraphics[width=0.95\linewidth]{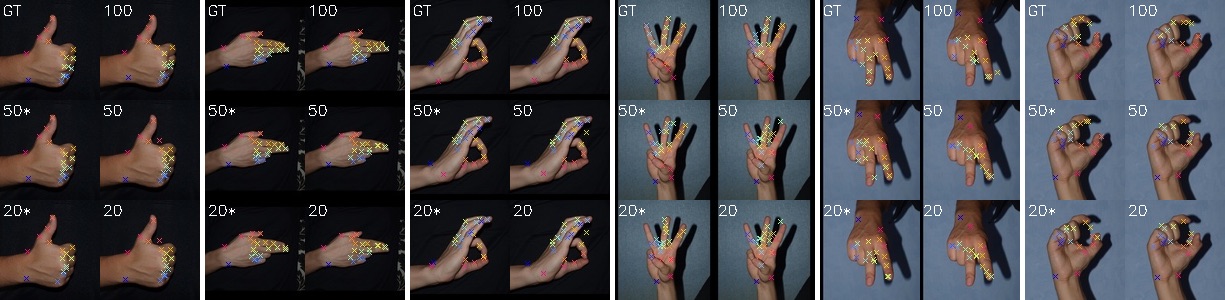}
\vspace{1.5em}
\end{tabular}
   \caption{Extra examples of our model predictions on the HGR1 \cite{kawulok2014PolishHands, nalepa2014PolishHands} test set. GT represents ground-trust annotations, while numbers 100, 50, and 20 indicate the percentage of the training set with labeled landmarks. Results are computed with Seq-MT (L+ELT+A) model (denoted *) and Seq-MT (L). Examples illustrate improvement of the landmark prediction by using the class label and the ELT cost in addition to the labeled landmarks. The last three examples on the bottom row show examples with high errors. Best viewed in color with zoom.}
\label{fig:hand_examples_2}
\end{center}
\end{figure*}

Figure \ref{fig:hand_examples_2} provides further landmark localization examples on hands dataset.

\subsection{Supplementary results on 300W dataset}
\begin{figure*}[ht]
\begin{center}
\includegraphics[width=.99\textwidth]{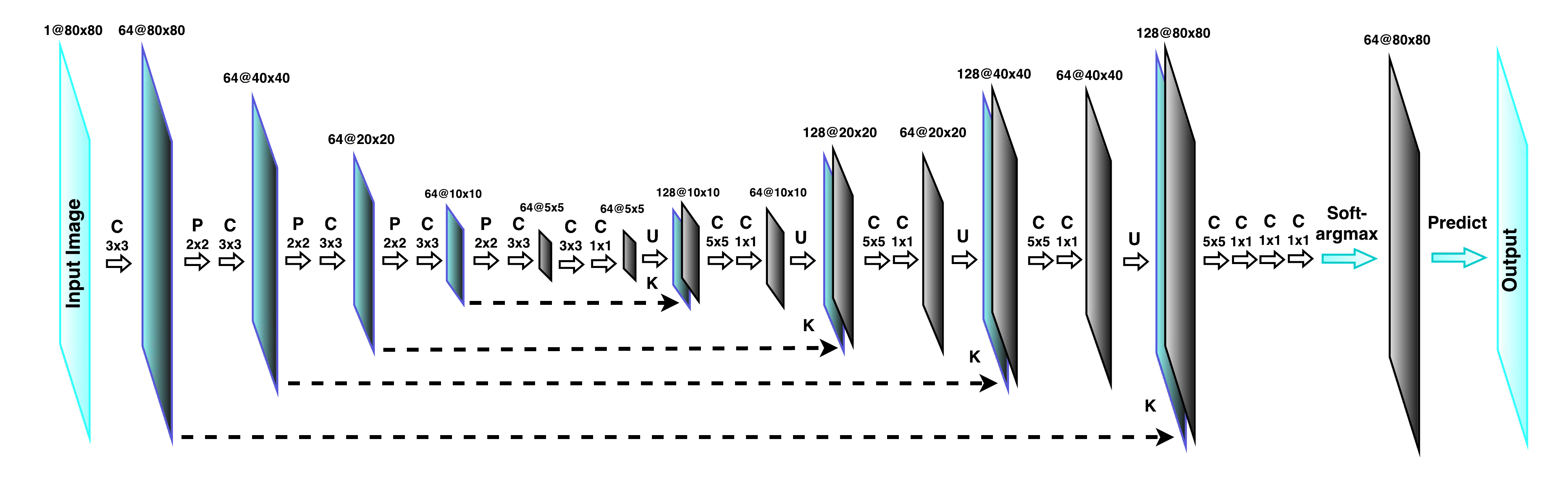}
\end{center}
   \caption{The ReCombinator Networks (RCN) \cite{honari2016recombinator} architecture used for experiments on 300W dataset. P indicates a pooling layer. All pooling layers have stride of 2. C indicates a convolutional layer. The number written below C indicates the convolution kernel size. All convolutions have stride of 1. U indicates an upsampling layer, where each feature map is upsampled to the next (bigger) feature map resolution. K indicates concatenation, where the upsampled features are concatenated with features of the same resolution before a pooling is applied to them. The dashed arrows indicate the feature maps are carried forward for concatenation. The solid arrows following each other, e.g. P, C, indicate the order of independent operations that are applied. The number written above feature maps in $n@w \times h$ format indicate number of feature maps $n$ and the width $w$ and height $h$ of the feature maps. On AFLW, we use 70 feature maps per layer (instead of 64) and we get two levels coarser to get to $1 \times 1$ resolution (instead of $5 \times 5$). On both datasets we shoud $\beta=100$ for soft-argmax layer.}
\label{fig:RCN}
\end{figure*}

\begin{figure*}[t]
\begin{center}
\rotatebox{90}{\begin{minipage}[b]{210pt}{
\begin{center}\tiny 
\hspace{.8mm} RCN $\stackanchor{+}{}$ \hspace{6mm}\hspace{4.8mm} RCN $\stackanchor{+}{}$ \hspace{8mm}\hspace{5mm} Seq-MT \hspace{6mm}\hspace{5mm} Seq-MT\\
\hspace{-2mm} (L+ELT) \hspace{6mm}\hspace{5.2mm} (L)  \hspace{8mm}\hspace{7.mm} (L+ELT)  \hspace{6mm}\hspace{6.5mm} (L) \\
\end{center}}\end{minipage}}\,
\includegraphics[width=.95\textwidth]{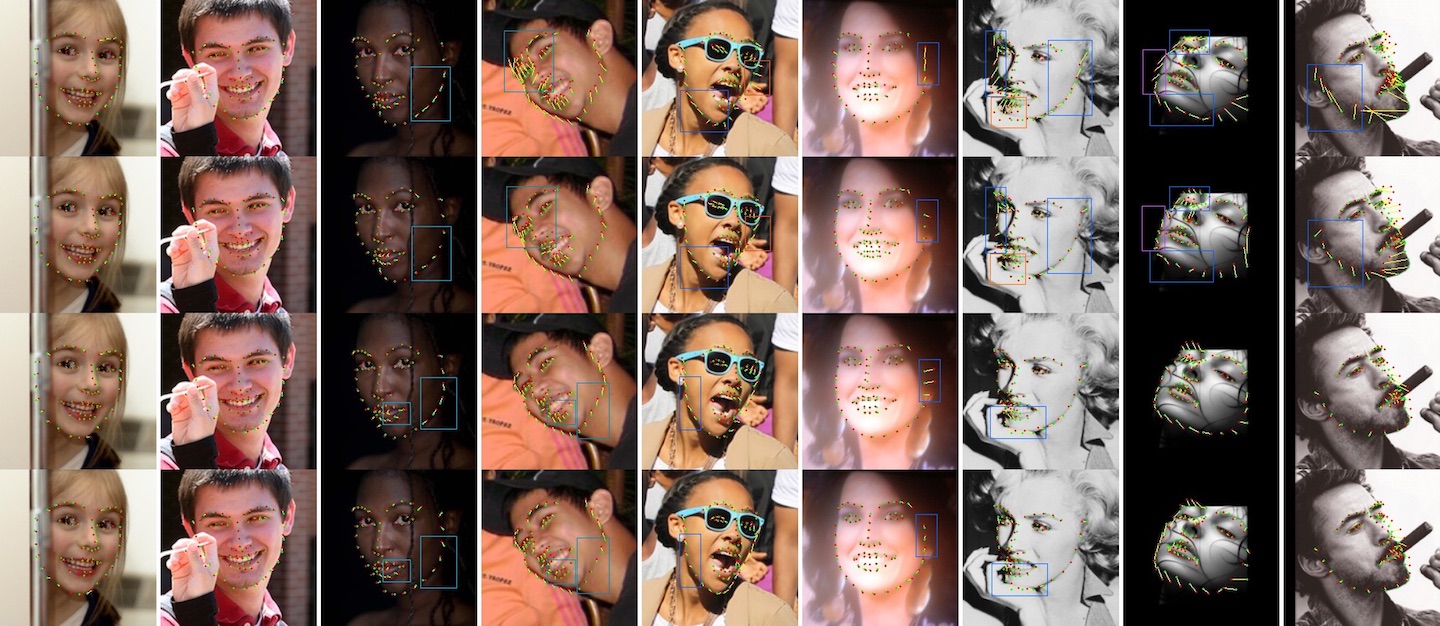}
\end{center}
   \caption{Extra examples of our model predictions on 300W \cite{sagonas2013300} test-set. The first two columns depict examples where all models get accurate predictions, The next 5 columns illustrate the improved accuracy obtained by using ELT loss in two different architectures (Seq-MT and RCN). The last two columns show difficult examples where error is high. The rectangles indicate the regions that landmarks are mostly affected. The green and red dots show ground truth (GT) and model predictions (MP), respectively. The yellow lines show the error by connecting GT and MP. Note that the ELT loss improves predictions in both architectures. Best viewed in color with zoom.
}
\label{fig:300W_examples_supp}
\end{figure*}

In Figure \ref{fig:RCN} we show the architecture of \textit{RCN $\stackanchor{+}{}$} 
used for 300W and AFLW datasets. In Figure \ref{fig:300W_examples_supp} we illustrate further samples from 300W dataset. The samples show the improved accuracy obtained in both \textit{Seq-MT} and \textit{RCN $\stackanchor{+}{}$} by using the ELT loss.

\subsection{Supplementary results on AFLW dataset}
In Table \ref{tab:aflw_pose} we show pose estimation error using different percentage of labelled data for RCN$\stackanchor{+}{}$ (L+ELT+A) model and compare the results to a model trained to estimate pose from GT landmarks. All models get close results compared to GT model indicating RCN$\stackanchor{+}{}$ (L+ELT+A) can do a reliable pose estimation using a small set of labelled landmarks. 

\begin{figure*}[ht]
\begin{center}
\begin{tabular}{c}
\rotatebox{90}{\begin{minipage}[b]{155pt}{
\begin{center}\tiny 
\shortstack{RCN$\stackanchor{+}{}$ \\ (L+ELT+A) \\ 100\%} \hspace{10mm} \shortstack{RCN$\stackanchor{+}{}$ \\ (L+ELT+A) \\ 1\%} 
\hspace{10mm} \shortstack{RCN$\stackanchor{+}{}$ \\ (L) \\ 1\%}
\end{center}}\end{minipage}}\,
\includegraphics[width=0.95\linewidth]{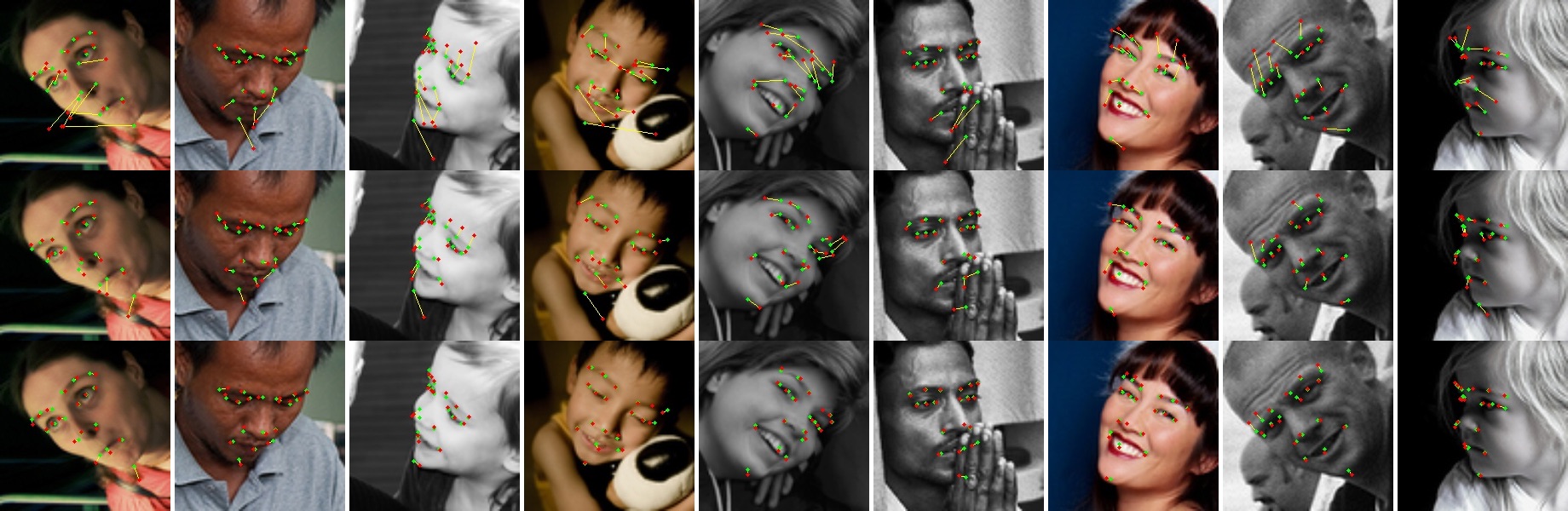}
\vspace{.5em}
\\
\rotatebox{90}{\begin{minipage}[b]{150pt}{
\begin{center}\tiny 
\shortstack{RCN$\stackanchor{+}{}$ \\ (L+ELT+A) \\ 100\%} \hspace{10mm} \shortstack{RCN$\stackanchor{+}{}$ \\ (L+ELT+A) \\ 1\%} 
\hspace{10mm} \shortstack{RCN$\stackanchor{+}{}$ \\ (L) \\ 1\%}
\end{center}}\end{minipage}}\,
\includegraphics[width=0.95\linewidth]{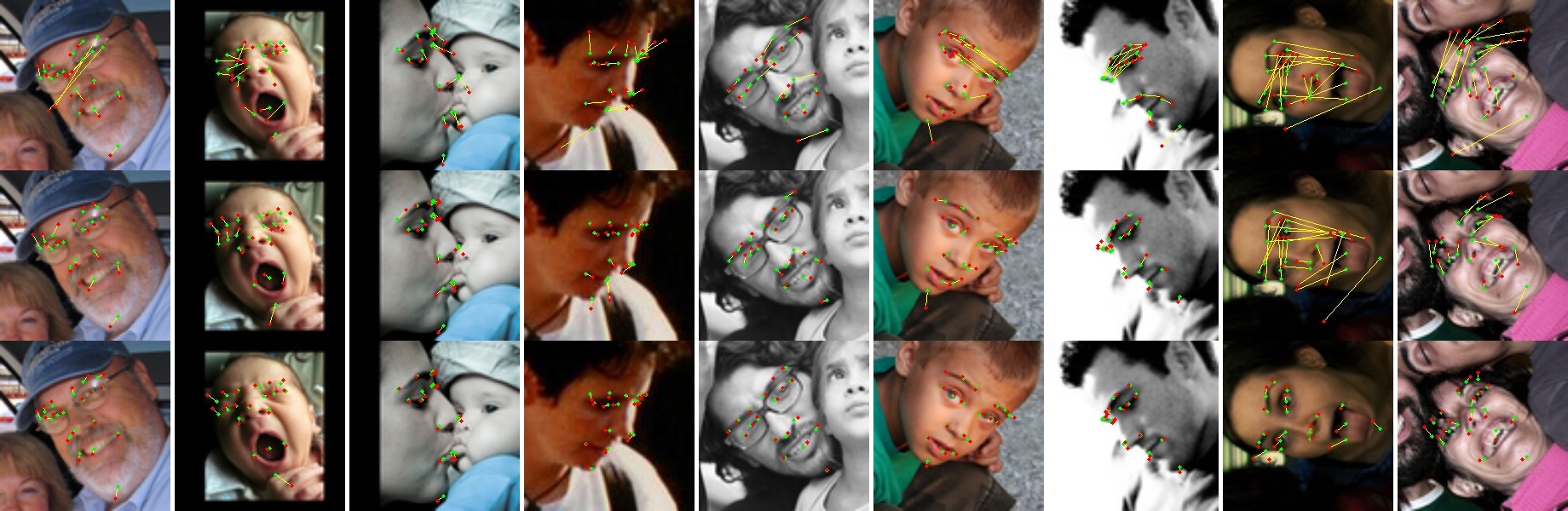}
\vspace{.5em}
\end{tabular}
   \caption{Extra examples of our model predictions on the AFLW test set. Comparing the first and second rows shows the improvement obtained by using ELT+A with only 1\% of labelled landmarks. Note the model trained using ELT+A preserves better the distribution over the landmarks. The last two columns in the bottom row show samples with high error on small percentage of labelled landmaks, which is due to extreme rotation. The bottom row shows the prediction using L+ELT+A on the entire set of labelled landmarks, which gets the best results. The green and red dots show ground truth (GT) and model predictions (MP), respectively. The yellow lines show the error by connecting GT and MP. Best viewed in color with zoom.}
\label{fig:aflw_examples_2}
\end{center}
\end{figure*}

Figure \ref{fig:aflw_examples_2} shows some samples on AFLW test set.

\begin{table}[t]
\caption{Emotion classification accuracy on Multi-PIE test set. In percent; higher is better.}
\label{tab:multiPIE_emotion}
\vskip2pt
\centering
\resizebox{1\linewidth}{!}{
\begin{tabular}{c*{5}{C}}
\toprule
 & \multicolumn{5}{c}{\textbf{Percentage of Images with Labeled Landmarks}} \\
\cmidrule{2-6}
\textbf{Model} & 5\% & 10\% & 20\% & 50\% & 100\% \\
\midrule
\midrule
Comm-MT (L+A) &  74.67 & 79.90 & 83.76 & 86.37 & 86.83 \\
Heatmap-MT (L+A) & \textbf{85.14} & \textbf{87.50} & \textbf{86.93} & \textbf{88.16} & \textbf{87.29} \\
\midrule
Seq-MT (L+A) &  78.78 & 82.62 & 84.69 & 84.03 & 84.86 \\
Seq-MT (L+A+ELT) &  82.90 & 84.57 & 84.85 & 86.48 & \\
\end{tabular}
}
\end{table}
\begin{table}[t]
\caption{Camera classification accuracy on Multi-PIE test set. In percent; higher is better.}
\label{tab:multiPIE_camera}
\vskip2pt
\centering
\resizebox{1\linewidth}{!}{
\begin{tabular}{c*{5}{C}}
\toprule
 & \multicolumn{5}{c}{\textbf{Percentage of Images with Labeled Landmarks}} \\
\cmidrule{2-6}
\textbf{Model} & 5\% & 10\% & 20\% & 50\% & 100\% \\
\midrule
Comm-MT (L+A) &  96.98 & 97.53 & 98.30 & 98.63 & 98.80 \\
Heatmap-MT (L+A) &  \textbf{98.46} & \textbf{98.99} & \textbf{98.99} &	\textbf{98.98} & \textbf{98.98} \\
\midrule
Seq-MT (L+A) &  97.97 & 98.31 &	98.50 &	98.96 & 98.92 \\
eq-MT (L+A+ELT) &  98.41 & 98.53 &	98.47 &	98.43 & \\
\end{tabular}
}
\end{table}
\begin{table}[t]
\caption{Classification accuracy on hands test set. In percent; higher is better.}
\vskip-2pt
\label{tab:hands_class}
\centering
\resizebox{1\linewidth}{!}{
\begin{tabular}{c*{5}{C}}
\toprule
 & \multicolumn{5}{c}{\textbf{Percentage of Images with Labeled Landmarks}} \\
\cmidrule{2-6}
\textbf{Model} & 5\% & 10\% & 20\% & 50\% & 100\% \\
\midrule
Comm-MT (L+A) &  60.86 & 69.64 & 69.20 & 76.03 & 73.42 \\
Heatmap-MT (L+A) & \textbf{83.74} & \textbf{87.86} & \textbf{87.55} & \textbf{90.29} & \textbf{89.27} \\
\midrule
Seq-MT (L+A) &  69.08 & 70.14 & 72.26 & 77.07 & 75.92 \\
Seq-MT (L+A+ELT) &  74.64 & 75.01 & 73.90 & 79.10 & \\
\end{tabular}
}
\end{table}
\begin{table}[t]
\caption{Pose degree estimation error on AFLW test set, as average of yaw, pitch, roll values. lower is better.}
\label{tab:aflw_pose}
\vskip-2pt
\centering
\resizebox{1\linewidth}{!}{
\begin{tabular}{c*{3}{D}}
\toprule
 & \multicolumn{3}{c}{\textbf{Percentage of Images with Labeled Landmarks}} \\
\cmidrule{2-4}
\textbf{Model} & 1\% & 5\% & 100\% \\
\midrule
RCN$\stackanchor{+}{}$(L+ELT+A) &  5.05 & 5.01 & 5.12 \\
\midrule
GT &  &  & 4.98 \\
\end{tabular}
}
\end{table}
 \begin{table}[t]
 \caption{Architecture details for Comm-MT Model on Blocks dataset.}
 \label{tab:arch_comm_MT}
 \vskip-2pt
 \centering
 \resizebox{.75\linewidth}{!}{
 \begin{tabular}{cc}
 \toprule
 \midrule
 \multicolumn{2}{c}{Input = $60 \times 60 \times 1$} \\
 \midrule
 \multicolumn{2}{c}{Conv $ 9 \times 9 \times 8$, ReLU, stride 1, SAME} \\
 \multicolumn{2}{c}{Conv $ 9 \times 9 \times 8$, ReLU, stride 1, SAME} \\
 \multicolumn{2}{c}{Conv $ 9 \times 9 \times 8$, ReLU, stride 1, SAME} \\ 
 \multicolumn{2}{c}{Conv $ 9 \times 9 \times 8$, ReLU, stride 1, SAME} \\
 \multicolumn{2}{c}{Conv $ 9 \times 9 \times 8$, ReLU, stride 1, SAME} \\
 \multicolumn{2}{c}{Pool $ 2 \times 2$, stride 2 } \\
 \multicolumn{2}{c}{Conv $ 9 \times 9 \times 8$, ReLU, stride 1, SAME} \\
 \multicolumn{2}{c}{Pool $ 2 \times 2$, stride 2 } \\
 \multicolumn{2}{c}{Conv $1 \times 1 \times 8$, ReLU, stride 1, SAME} \\
 \multicolumn{2}{c}{Conv $1 \times 1 \times 8$, ReLU, stride 1, SAME} \\
 \multicolumn{2}{c}{FC $\#units=256$, ReLU, dropout-prob=.25} \\
 \multicolumn{2}{c}{FC $\#units=256$, ReLU, dropout-prob=.25} \\
 \midrule
 Classification branch & Landmark localization branch \\
 \midrule
 FC $\#units=15$, Linear & FC $\#units=10$, Linear \\
softmax(dim=15) & \\
\midrule
\end{tabular}
}
\end{table}
\begin{table}[t]
 \caption{Architecture details for Heatmap-MT Model on Blocks datasets.}
 \label{tab:arch_heatmap_MT}
 \vskip-2pt
 \centering
 \resizebox{1\linewidth}{!}{
 \begin{tabular}{cc}
 \toprule
 \midrule
 \multicolumn{2}{c}{Input = $60 \times 60 \times 1$} \\
 \midrule
 \multicolumn{2}{c}{Conv $ 9 \times 9 \times 8$, ReLU, stride 1, SAME} \\
 \multicolumn{2}{c}{Conv $ 9 \times 9 \times 8$, ReLU, stride 1, SAME} \\
 \multicolumn{2}{c}{Conv $ 9 \times 9 \times 8$, ReLU, stride 1, SAME} \\  
 \multicolumn{2}{c}{Conv $ 9 \times 9 \times 8$, ReLU, stride 1, SAME} \\
 \multicolumn{2}{c}{Conv $ 9 \times 9 \times 8$, ReLU, stride 1, SAME} \\  
 \multicolumn{2}{c}{Conv $ 9 \times 9 \times 8$, ReLU, stride 1, SAME} \\
 \multicolumn{2}{c}{Conv $1 \times 1 \times 8$, ReLU, stride 1, SAME} \\
 \multicolumn{2}{c}{Conv $1 \times 1 \times 5$, ReLU, stride 1, SAME} \\
 \midrule
 classification branch & landmark localization branch \\
 \midrule
 Pool $ 2 \times 2$, stride 2 & --- \\
 Conv $ 9 \times 9 \times 8$, ReLU, stride 1, SAME &  --- \\
 Pool $ 2 \times 2$, stride 2 &  --- \\
 Conv $ 9 \times 9 \times 8$, ReLU, stride 1, SAME&  --- \\
 Pool $ 2 \times 2$, stride 2 &  --- \\
 Conv $ 9 \times 9 \times 8$, ReLU, stride 1, SAME&  --- \\
 Pool $ 2 \times 2$, stride 2 &  --- \\
 Conv $ 9 \times 9 \times 8$, ReLU, stride 1, SAME&  --- \\
 FC $\#units=256$, ReLU, dropout-prob=.25 &  --- \\
 FC $\#units=256$, ReLU, dropout-prob=.25 &  --- \\
 FC $\#units=15$, Linear &  --- \\
softmax(dim=15) & softmax(dim=$60 \times 60$) \\
\midrule
\end{tabular}
}
\end{table}
\subsection{Architecture details}
\label{sec:arch_details}
The architecture details of Seq-MT model on different datasets can be seen in Tables \ref{tab:arch_seq}, \ref{tab:arch_seq_real_1} and \ref{tab:arch_seq_real_2}. Architecture details of Comm-MT and Heatmap-MT for Blocks dataset are shown in Tables \ref{tab:arch_comm_MT} and \ref{tab:arch_heatmap_MT}. For other dataset, the kernel size and the number of feature maps for conv layers and the number of units for FC layers change similar to Seq-MT model on those datasets.
\begin{table*}[t]
\caption{Architecture details of Seq-MT model used for Shapes and Blocks datasets.
Each conv layer has three values as $w \times h \times n$ indicating width (w), height (h) of kernel and the number of feature maps (n) of the convolutional layer. SAME indicates the input map is padded with zeros such that input and output maps have the same resolution.}
\label{tab:arch_seq}
\vskip2pt
\centering
\resizebox{.55\linewidth}{!}{
\begin{tabular}{c|c}
\toprule
\textbf{Shapes Dataset} & \textbf{Blocks Dataset} \\
\midrule
Model HP: $\lambda = 0, \alpha=0, \gamma=0, \beta=1$, ADAM & Model HP: $\lambda = 1, \alpha=1, \beta=1$, ADAM \\
\midrule
Landmark Localization Network & Landmark Localization Network\\
\midrule
Input = $60 \times 60 \times 1$  & Input = $60 \times 60 \times 1$ \\
Conv $7 \times 7 \times 16$, ReLU, stride 1, SAME & Conv $ 9 \times 9 \times 8$, ReLU, stride 1, SAME  \\
Conv $7 \times 7 \times 16$, ReLU, stride 1, SAME & Conv $ 9 \times 9 \times 8$, ReLU, stride 1, SAME \\
Conv $7 \times 7 \times 16$, ReLU, stride 1, SAME & Conv $ 9 \times 9 \times 8$, ReLU, stride 1, SAME \\
Conv $7 \times 7 \times 16$, ReLU, stride 1, SAME & Conv $ 9 \times 9 \times 8$, ReLU, stride 1, SAME \\
Conv $7 \times 7 \times 16$, ReLU, stride 1, SAME & Conv $ 9 \times 9 \times 8$, ReLU, stride 1, SAME \\
Conv $7 \times 7 \times 16$, ReLU, stride 1, SAME & Conv $ 9 \times 9 \times 8$, ReLU, stride 1, SAME \\
Conv $1 \times 1 \times 16$, ReLU, stride 1, SAME & Conv $1 \times 1 \times 8$, ReLU, stride 1, SAME \\
Conv $1 \times 1 \times 2$, ReLU, stride 1, SAME & Conv $1 \times 1 \times 5$, ReLU, stride 1, SAME\\
soft-argmax(num\_channels=2) & soft-argmax(num\_channels=5) \\ 
\midrule
Classification Network   & Classification Network \\
\midrule
FC $\#units=40$, ReLU   & FC $\#units=256$, ReLU, dropout-prob=.25 \\
FC $\#units=2$, Linear & FC $\#units=256$, ReLU, dropout-prob=.25\\
						 & FC $\#units=15$, Linear \\
softmax(dim=2)           & softmax(dim=15) \\
\midrule
\end{tabular}
}
\end{table*}
\begin{table*}[t]
\caption{Architecture details of Seq-MT model used for Hands and Multi-PIE datasets.}
\label{tab:arch_seq_real_1}
\vskip2pt
\centering
\resizebox{.9\linewidth}{!}{
\begin{tabular}{c|cc}
\toprule
\textbf{Hands Dataset} & \multicolumn{2}{c}{\textbf{Multi-PIE Dataset}} \\
\midrule
Model HP: $\lambda=0.5, \alpha=0.3, \gamma=10^{-5}, \beta=0.001$, ADAM & \multicolumn{2}{c}{Model HP: $\lambda = 2, \alpha=0.3, \gamma=10^{-5}, \beta=0.001$, ADAM} \\
\midrule
\multicolumn{3}{c}{\textbf{Preprocessing:} scale and translation [-10\%, 10\%] of face bounding box, rotation [-20, 20] applied randomly to every epoch.} \\
\midrule
Landmark Localization Network & \multicolumn{2}{c}{Landmark Localization Network} \\
\midrule
Input = $64 \times 64 \times 1$  & \multicolumn{2}{c}{Input = $64 \times 64 \times 1$} \\
Conv $9 \times 9 \times 64$, ReLU, stride 1, SAME & \multicolumn{2}{c}{Conv $ 9 \times 9 \times 64$, ReLU, stride 1, SAME} \\
Conv $9 \times 9 \times 64$, ReLU, stride 1, SAME & \multicolumn{2}{c}{Conv $ 9 \times 9 \times 64$, ReLU, stride 1, SAME} \\
Conv $9 \times 9 \times 64$, ReLU, stride 1, SAME & \multicolumn{2}{c}{Conv $ 9 \times 9 \times 64$, ReLU, stride 1, SAME} \\
Conv $9 \times 9 \times 64$, ReLU, stride 1, SAME & \multicolumn{2}{c}{Conv $ 9 \times 9 \times 64$, ReLU, stride 1, SAME} \\
Conv $9 \times 9 \times 64$, ReLU, stride 1, SAME & \multicolumn{2}{c}{Conv $ 9 \times 9 \times 64$, ReLU, stride 1, SAME} \\
Conv $9 \times 9 \times 25$, ReLU, stride 1, SAME & \multicolumn{2}{c}{Conv $ 9 \times 9 \times 68$, ReLU, stride 1, SAME} \\
soft-argmax(num\_channels=25)                     & \multicolumn{2}{c}{soft-argmax(num\_channels=68)} \\ 
\midrule
Classification Network 						      & Emotion Classification Branch & Camera Classification Branch \\
\midrule
FC $\#units=256$, ReLU, dropout-prob=.5		  & FC $\#units=256$, ReLU, dropout-prob=.25 & FC $\#units=256$, ReLU, dropout-prob=.25\\
FC $\#units=256$, ReLU, dropout-prob=.5		  & FC $\#units=256$, ReLU, dropout-prob=.25 & FC $\#units=256$, ReLU, dropout-prob=.25\\
FC $\#units=27$, Linear      				  & FC $\#units=6$, Linear & FC $\#units=5$, Linear \\
softmax(dim=27) 								  & softmax(dim=6)            & softmax(dim=5) \\
\midrule
\end{tabular}
}
\end{table*}
\begin{table*}[t]
\caption{Architecture details of Seq-MT model used for 300W datasets.}
\label{tab:arch_seq_real_2}
\vskip2pt
\centering
\resizebox{0.5\linewidth}{!}{
\begin{tabular}{c}
\toprule
\textbf{300W Dataset} \\
\midrule
Model HP: $\lambda=2.0, \alpha=2.0, \gamma=10^{-5}, \beta=0.01$, ADAM \\
\midrule
\shortstack{\textbf{Preprocessing:} scale and translation [-10\%, 10\%] of face bounding box, \\
rotation [-30, 30] applied randomly to every epoch.} \\
\midrule
Landmark Localization Network \\
\midrule
Input = $64 \times 64 \times 1$  				  \\
Conv $9 \times 9 \times 32$, ReLU, stride 1, SAME \\
Conv $9 \times 9 \times 32$, ReLU, stride 1, SAME \\
Conv $9 \times 9 \times 32$, ReLU, stride 1, SAME \\
Conv $9 \times 9 \times 32$, ReLU, stride 1, SAME \\
Conv $9 \times 9 \times 32$, ReLU, stride 1, SAME \\
Conv $9 \times 9 \times 32$, ReLU, stride 1, SAME \\
Conv $9 \times 9 \times 32$, ReLU, stride 1, SAME \\
Conv $9 \times 9 \times 68$, ReLU, stride 1, SAME \\
soft-argmax(num\_channels=68)                     \\ 
\midrule
\end{tabular}
}
\end{table*}

\end{document}